\documentclass{article} 
\usepackage{iclr2027_conference,times}

\usepackage{amsmath,amsfonts,bm}

\def\eqref#1{equation~\ref{#1}}

\def\1{\bm{1}}

\DeclareMathAlphabet{\mathsfit}{\encodingdefault}{\sfdefault}{m}{sl}
\SetMathAlphabet{\mathsfit}{bold}{\encodingdefault}{\sfdefault}{bx}{n}

\usepackage{hyperref}
\usepackage{url}
\usepackage{subcaption}
\usepackage{amssymb}
\usepackage{amsmath}
\usepackage{booktabs}
\usepackage{graphicx}
\usepackage{placeins}
\usepackage{xcolor}

\usepackage{listings}
\lstdefinestyle{python}{
  language=Python,
  basicstyle=\ttfamily\small,
  keywordstyle=\color{blue!70!black}\bfseries,
  commentstyle=\color{green!40!black}\itshape,
  stringstyle=\color{orange!70!black},
  backgroundcolor=\color{gray!5},
  breaklines=true,
  columns=fullflexible,
  keepspaces=true,
  showstringspaces=false,
  tabsize=4
}

\definecolor{codebackground}{HTML}{F8FAFC}
\definecolor{codeborder}{HTML}{CBD5E1}

\title{Latent Compaction with Test-Time Training}

\author{Yingfa Chen, Xingyu Shen, Chaojun Xiao, Xu Han \\
NLP Group, DCST, IAI, BNRIST, Tsinghua University, Beijing, China \\
\texttt{chenyingfa1999@gmail.com}
}

\usepackage{xspace}
\usepackage[most]{tcolorbox}
\newtcolorbox{hypothesisbox}{colback=blue!3,colframe=blue!60!black,boxrule=0.8pt,arc=2pt,left=6pt,right=6pt,top=6pt,bottom=6pt}

\iclrfinalcopy 
\begin{document}

\maketitle

\begin{abstract}
Long contexts enlarge key--value (KV) caches, increasing memory use and attention costs during generation. Compressing these caches must preserve information for unseen questions. We introduce \textbf{Latent Compaction}, which learns a compact document-specific cache through test-time training. We initialize cache entries using reconstruction importance, teacher attention, and key-space diversity. Training first fits attention outputs and log-normalizers to those of the full cache, with reconstruction-selected entries held fixed. It then jointly optimizes the entire cache through end-to-end distillation from the full-cache model. Model weights remain frozen throughout, and training uses only generic prompts about the document, without access to downstream questions or answers. Experiments on RULER, LongBench, and Phonebook Lookup demonstrate substantial gains over budget-matched compression baselines. On a RULER, for example, our method achieves 96.7\% accuracy at $32\times$ compression, compared with 40.0\% for KVZip+. The method also transfers across model scales and families, with retrieval gains on both Qwen3 and MiniCPM5. These results show that treating the KV cache as a neural network and optimizing it with test-time training can preserve retrieval accuracy under aggressive compression.\footnote{The code is released at \url{https://www.github.com/THUNLP/latent-compaction}}

\end{abstract}

\section{Introduction}
\label{sec:introduction}

Large key--value (KV) caches are a major bottleneck to efficient LLM inference: long contexts limit batch sizes and increase memory-bandwidth demands at each decoding step~\citep{zhang2023h2o,liu2024kivi}, motivating context compression.

Frontier-model platforms already offer \emph{context compaction} to sustain conversations beyond a single context window~\citep{anthropic2026compaction,openai2026compaction}. Claude, for example, replaces older context with a text summary. Summaries can omit details whose importance only becomes clear later~\citep{anthropic2025contextengineering}, motivating compression of the model's KV representation directly. We study document compression before downstream questions arrive: the cache must preserve information for unseen queries, while preparation costs can be amortized across repeated use.

Eviction methods select original KV entries using attention or reconstruction~\citep{li2024snapkv,cai2024pyramidkv,kim2025kvzip}. Cartridges learns compact caches through self-study~\citep{eyuboglu2025cartridges}, while Attention Matching fits attention outputs and mass~\citep{zweiger2026attentionmatching}. These approaches motivate test-time training of document memory~\citep{sun2020testtimetraining}: \emph{can gradient descent on a compact KV cache preserve document information under an aggressive memory budget?}

We introduce \textbf{Latent Compaction}, which trains a document-specific KV cache with frozen model weights. Initialization combines globally selected preservers with anchors and diversifiers. Stage 1 fixes preservers and fits the other entries to full-cache attention outputs and log-normalizers. Stage 2 jointly optimizes all entries using the model's predictions on generic document prompts. Downstream questions and answers are excluded; the trained cache is fixed during reuse. For a 4K document at $32\times$ compression, each layer/head averages 96 preservers and 32 anchors and diversifiers.

Experiments on RULER, LongBench, and Phonebook Lookup demonstrate the benefits of Latent Compaction across synthetic retrieval and natural-document understanding tasks. The method substantially improves retrieval accuracy over budget-matched compression baselines across context lengths and compression ratios, while LongBench results show gains on broader long-context tasks. Additional Phonebook Lookup experiments demonstrate transfer across model scales and families: the retrieval advantages extend to multiple Qwen3 model sizes and MiniCPM5 models. Ablations identify the roles of initialization, both training stages, and attention normalization. These results concern logical KV-pair budgets rather than measured memory savings or inference speedups.

Our contributions are:
\begin{itemize}
    \item A novel method for learning compact context representation trained with gradient descent and initialized through complementary selection rules.
    \item A document-specific test-time training procedure that learns from the frozen model's attention and predictions without access to downstream questions or answers.
    \item Superior performance on a wide range of empirical downstream evaluation across three benchmarks and multiple model scales and families, with ablations of initialization, training, and attention normalization.
\end{itemize}

\begin{figure}
    \centering
    \includegraphics[width=0.99\linewidth]{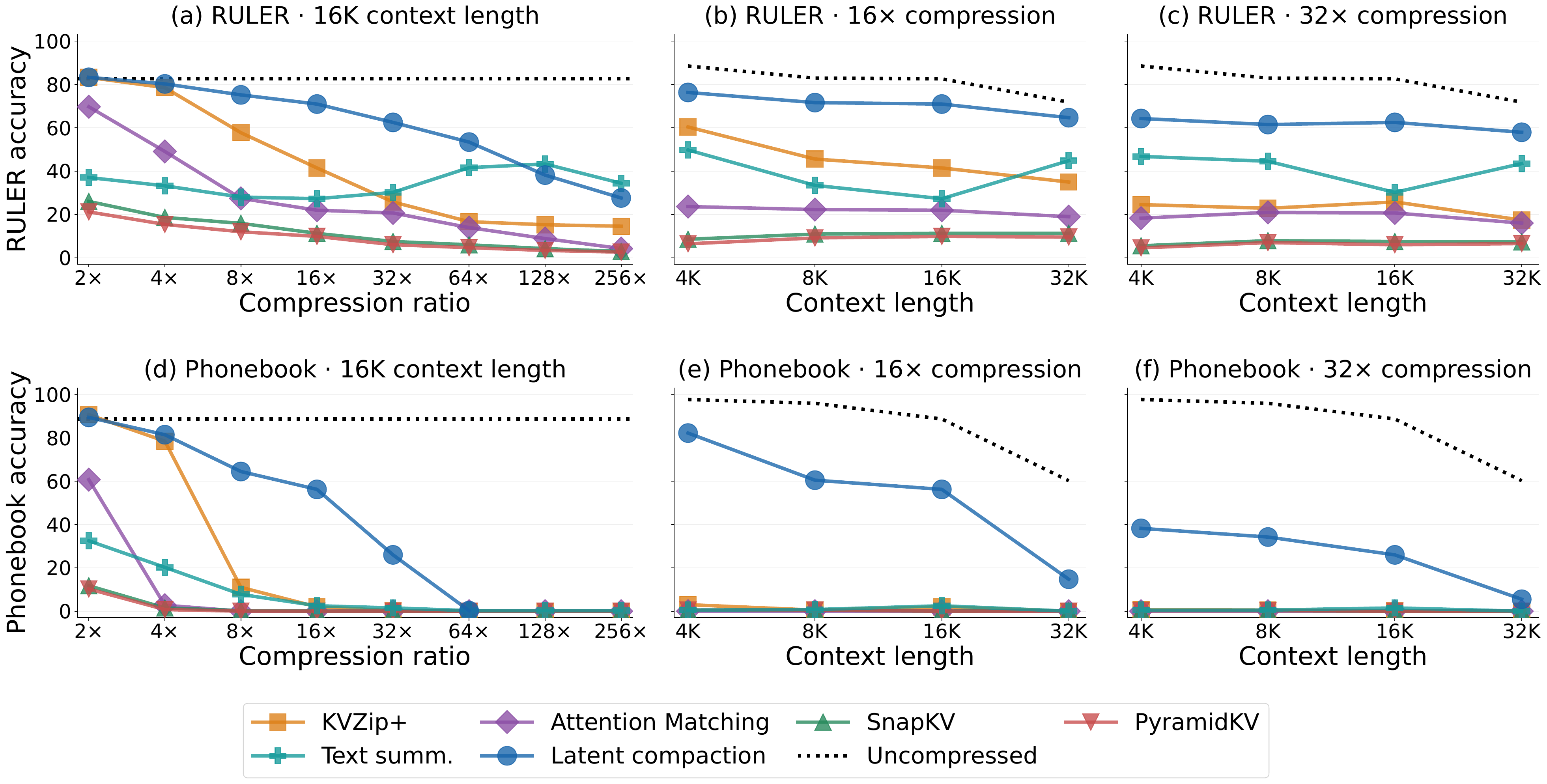}
    \caption{Latent Compaction and baselines across compression ratios and context lengths on RULER and Phonebook Lookup with Qwen3-1.7B.}
    \label{fig:main-results}
\end{figure}

\section{Related Work}
\label{sec:related-work}

\paragraph{KV-Cache Eviction and Budget Allocation.}
Eviction methods retain selected original KV pairs. H$_2$O combines recent and high-attention tokens \citep{zhang2023h2o}, while SnapKV selects positions using an observation window \citep{li2024snapkv}. PyramidKV varies budgets across layers \citep{cai2024pyramidkv}, and Ada-KV adapts budgets across heads \citep{feng2024adakv}. KVZip scores each entry by its maximum attention weight during context reconstruction~\citep{kim2025kvzip}. KVZip+, introduced by \citet{jegou2026kvzap}, rescales each weight by the projected value norm divided by the input hidden-state norm before taking the maximum. This accounts for each value's relative contribution to the residual stream. Their KVzap method learns lightweight predictors of these scores to accelerate pruning. We use KVZip+ scores to initialize preservers before cache optimization. We evaluate query-agnostic SnapKV and PyramidKV adaptations that withhold downstream questions.

\paragraph{Learned Context Representations.}
Prefix-tuning optimizes continuous prefixes with frozen model weights \citep{li2021prefixtuning}. AutoCompressors and the In-context Autoencoder learn compression mechanisms across examples, producing compact summary vectors or memory slots \citep{chevalier2023autocompressors,ge2023icae}. Cartridges instead optimizes a compact KV cache for an individual corpus through self-study and context distillation \citep{eyuboglu2025cartridges}. Attention Matching fits full-cache attention outputs and mass through key selection, biases, and values \citep{zweiger2026attentionmatching}. Our related stage-1 loss is followed by end-to-end refinement of all entries. RestoreKV generates restoration entries using a shared LoRA mechanism \citep{baek2026restorekv}; we optimize document-specific slots directly.

\paragraph{Test-Time Training.}
Test-time training~(TTT) adapts model parameters using a self-supervised objective on an unlabeled test input before prediction \citep{sun2020testtimetraining}. For language models, \citet{hardt2024tttnn} fine-tune model weights on retrieved nearest-neighbor text for each test input. TTT layers make an online-learned model the recurrent hidden state \citep{sun2025tttlayers}, while TTT-E2E updates Transformer weights on the input context through next-token prediction, with a meta-learned initialization \citep{tandon2025ttte2e}. Latent Compaction instead updates document-specific KV slots in an existing frozen Transformer, using full-cache teacher attention and predictions. It prepares the compact cache before downstream questions are known and reuses that cache across questions without further optimization.

\begin{figure*}[!t]
    \centering
    \includegraphics[width=0.99\linewidth]{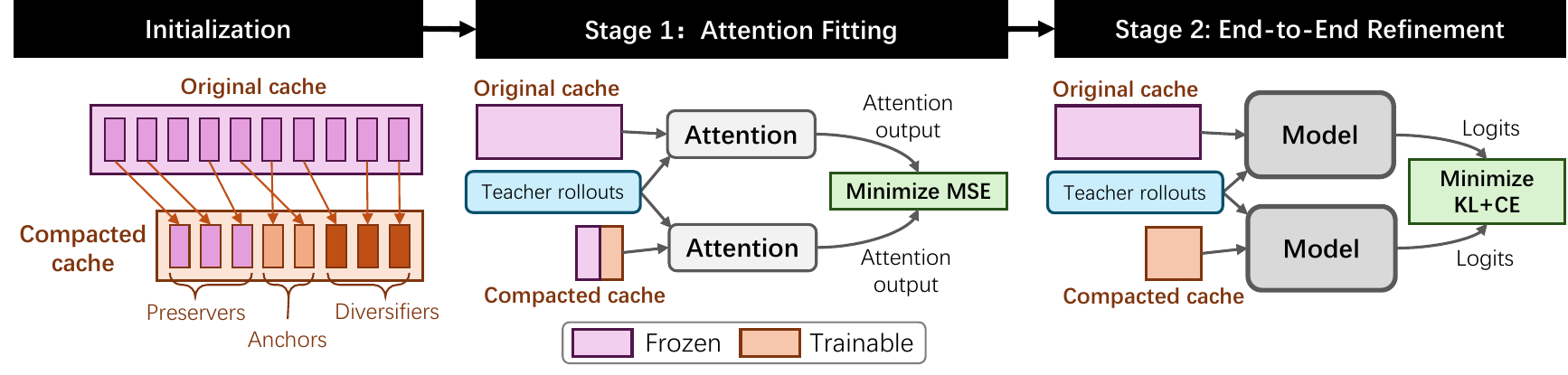}
    \caption{Latent Compaction training pipeline. Preservers, anchors, and diversifiers are initialized from the original cache. Stage 1 freezes preservers and fits anchors and diversifiers to full-cache attention outputs. Stage 2 refines all entries through end-to-end distillation.}
    \label{fig:method}
\end{figure*}

\section{Method: Latent Compaction}
\label{sec:method}

Latent Compaction constructs a compact, trainable KV cache for a document before downstream questions arrive. Model weights remain frozen throughout, and the optimized cache is reused across questions. 

In what follows, Section~\ref{sec:compressed-memory} defines the memory architecture and its interaction with subsequent tokens. Section~\ref{sec:initialization} explains the preserver, anchor, and diversifier initialization and budget allocation. Sections~\ref{sec:training-stage-1} and~\ref{sec:training-stage-2} describe attention fitting and end-to-end refinement, respectively. Section~\ref{sec:teacher-rollout} details the generic teacher rollouts that provide supervision for both stages.

\paragraph{Preliminaries: Attention and KV Caching.}
Consider a query head and its associated KV head in one Transformer layer. Let $q,k_i,v_i\in\mathbb R^d$ denote the query and the key and value at cache entry $i$, where $d$ is the head dimension. Let $\mathcal I$ index entries causally visible to $q$: document entries plus any question or generated-token entries at or before its position. The attention output $A(q)\in\mathbb R^d$ and scalar normalizer $Z(q)>0$ are
\begin{equation}
    A(q)=\frac{\sum_{i\in\mathcal I}\exp\left(q^\top k_i / \sqrt{d} \right)v_i}{Z(q)},
    \qquad
    Z(q)=\sum_{i\in\mathcal I} \exp \left(q^\top k_i / \sqrt{d}\right),
    \label{eq:standard-attention}
\end{equation}
Queries and keys include positional encoding. An $N$-token document stores $2LHNd$ key/value scalars across $L$ layers and $H$ KV heads; grouped-query attention shares each KV head among multiple query heads. We compress this document cache for reuse across questions. Below, $s=1/\sqrt d$ is the attention scale.

\subsection{Memory Architecture}
\label{sec:attention-preliminaries}
\label{sec:compressed-memory}

Given an average budget of $B$ KV pairs per layer and KV head, we initialize $LHB$ KV slots in total. The number of slots may vary across layer--head pairs. Each slot also carries a scalar attention bias to compensate for the change in attention mass caused by compression. 

\paragraph{Attention Biases.}
For one layer and KV head, let $n$ be its number of compacted document entries. Omitting layer/head indices, let
$K_{\mathrm{cmp}},V_{\mathrm{cmp}}\in\mathbb R^{n\times d}$
contain their keys and values as rows, and let
$b_{\mathrm{cmp}}\in\mathbb R^n$ contain their biases, initialized to zero.
For a query head associated with this KV head, $q\in\mathbb R^d$. Following Attention Matching \citep{zweiger2026attentionmatching}, biases adjust attention mass:
\begin{align}
    \ell(q)&=sK_{\mathrm{cmp}}q+b_{\mathrm{cmp}},\\
    A_{\mathrm{cmp}}(q)
    &=V_{\mathrm{cmp}}^\top\operatorname{softmax}(\ell(q)).
    \label{eq:compressed-attention}
\end{align}
The score vector $\ell(q)\in\mathbb R^n$ and its softmax weights have one component per compacted entry; $A_{\mathrm{cmp}}(q)\in\mathbb R^d$ is the resulting per-head attention output.
A bias $b_j\in\mathbb R$ multiplies entry $j$'s unnormalized weight by $e^{b_j}$, adjusting both its relative contribution and the document's total attention mass. Cache parameters are learned per document and fixed during downstream inference.

\paragraph{Handling Subsequent Tokens.}
Question and generated tokens append ordinary, unbiased KV pairs directly after the document, outside its budget and without renumbering positions. For a query $q_t\in\mathbb R^d$ at token position $t$, define the scalar compressed-cache attention normalizer
\begin{equation}
    Z_{\mathrm{cmp}}(q_t)=\mathbf 1_n^\top\exp(\ell(q_t)).
    \label{eq:compressed-normalizer}
\end{equation}
Here $\mathbf 1_n\in\mathbb R^n$ is the all-ones vector and exponentiation is elementwise. Let $A_{\mathrm{new},t}\in\mathbb R^d$ and $Z_{\mathrm{new},t}\geq0$ be the attention output and scalar normalizer over causally accessible question and generated-token entries for $q_t$ (taking both as zero if there are none). The combined output $A_t\in\mathbb R^d$ is
\begin{equation}
    A_t=
    \frac{Z_{\mathrm{cmp}}(q_t)A_{\mathrm{cmp}}(q_t)
    +Z_{\mathrm{new},t}A_{\mathrm{new},t}}
    {Z_{\mathrm{cmp}}(q_t)+Z_{\mathrm{new},t}}.
    \label{eq:autoregressive-attention}
\end{equation}
This equals a single softmax over compressed and subsequent entries, with learned biases applied only to the former. Normalizers and mixture weights are computed in log-space.

\subsection{Initialization}
\label{sec:initialization}

We initialize compact caches from original KV pairs using three complementary kinds of entries:

\begin{itemize}
    \item \textbf{Preserver.}
    We rank the KVZip+ reconstruction scores of all original KV pairs from every layer and KV head together~\citep{kim2025kvzip,jegou2026kvzap}. The highest-scoring pairs fill one total preserver budget, so layer--head pairs can receive different counts.

    \item \textbf{Anchor.}
    One quarter of the anchor--diversifier budget uses original pairs with high normalized teacher attention. We also include the first document position as a potential attention sink~\citep{xiao2024efficientstreaminglanguagemodels}.

    \item \textbf{Diversifier.}
    In each layer and KV head, choose each new key farthest from its nearest anchor or previously selected diversifier. Preservers are excluded from the distance reference set.
\end{itemize}

Preservers are frozen in stage 1; all entries train in stage 2. Anchors and diversifiers also train in stage 1. The rules combine high-scoring original entries with attention-informed seeds and broad key-space coverage.

\paragraph{Budget Allocation}

With an average budget of $B$ entries per layer and KV head, we reserve $M=B/4$ per head for anchors and diversifiers. The other $LH(B-M)$ slots come from one ranking of KVZip+ scores across all layers, KV heads, and document positions. One quarter of each head's $M$ slots are \emph{anchors}: the first document entry and the highest-scoring remaining entries under maximum teacher attention,
\begin{equation}
    a_i=\max_{q\in\mathcal Q}
    \left[\operatorname{softmax}(sKq)\right]_i,
\end{equation}
where $i\in\{1,\ldots,N\}$ indexes the original document positions, $K\in\mathbb R^{N\times d}$ contains their keys for the corresponding layer and KV head, $s=1/\sqrt d$, and $\mathcal Q\subset\mathbb R^d$ contains teacher queries from the associated query heads (Section~\ref{sec:teacher-rollout}). Thus $a_i$ is a scalar score for document position $i$. The first entry counts as an anchor.

The remaining entries are \emph{diversifiers}, which are selected by farthest-first traversal \citep{gonzalez1985clustering}. Starting with the set of anchor indices $S\subseteq\{1,\ldots,N\}$, we repeatedly add
\begin{equation}
    i^\star=\arg\max_{i\notin S}
    \min_{j\in S}\|k_i-k_j\|_2^2
\end{equation}
until $|S|=M$, where $k_i$ is row $i$ of $K$. The set $S$ starts with anchors and grows only with diversifiers, never with preserver selections. Selected keys keep their original values; stage 2 optimizes all entries.

\subsection{Training Stage 1: Attention Fitting}
\label{sec:training-stage-1}

We fit each layer independently using captured teacher queries. The attention outputs and log-normalizers are computed over the complete document cache, including the frozen preservers:
\begin{equation}
    \mathcal L_{\mathrm{attn}}
    =
    \operatorname{MSE}
    \left(\widehat A_{\mathrm{cmp}},\widehat A_{\mathrm{full}} \right)
    +
    \operatorname{MSE}
    \left(\widehat{\log Z}_{\mathrm{cmp}},
     \widehat{\log Z}_{\mathrm{full}} \right).
    \label{eq:attention-fitting-loss}
\end{equation}
For each captured query, $A_{\mathrm{cmp}},A_{\mathrm{full}}\in\mathbb R^d$ are compressed- and full-document attention outputs, while $\log Z_{\mathrm{cmp}},\log Z_{\mathrm{full}}\in\mathbb R$ are their log-normalizers. MSE averages over captured queries and the relevant output dimensions. Hats denote standardization using fixed teacher statistics per query head: output statistics are computed over query positions and channels, and log-normalizer statistics over query positions. Standard deviations are floored at $10^{-4}$. Matching log-normalizers calibrates document attention relative to subsequent tokens. Anchor and diversifier keys, values, and biases receive updates; preserver keys, values, and biases remain frozen in stage 1.

\subsection{Training Stage 2: End-to-End Refinement}
\label{sec:training-stage-2}

We refine all cache entries jointly. At step $t\in\{1,\ldots,T\}$ of a length-$T$ teacher continuation, $p_t,r_t\in[0,1]^{|\mathcal V|}$ are the full-cache and compressed-cache probability vectors over vocabulary $\mathcal V$ under the same text prefix; $p_{t,v}$ and $r_{t,v}$ denote their components for token $v$. We retain the teacher's top-20 token set $\mathcal T_t\subset\mathcal V$ individually and group the rest into a \emph{tail} category with probabilities $p_{t,\mathrm{tail}}=1-\sum_{v\in\mathcal T_t}p_{t,v}$ and $r_{t,\mathrm{tail}}=1-\sum_{v\in\mathcal T_t}r_{t,v}$. Both models use the teacher's token set, without renormalizing the top-20 probabilities. We minimize
\begin{align}
    \mathcal L_{\mathrm{KL}}
    &=\frac{1}{T}\sum_{t=1}^{T}\left[
    \sum_{v\in\mathcal T_t}p_{t,v}\log\frac{p_{t,v}}{r_{t,v}}
    +p_{t,\mathrm{tail}}\log\frac{p_{t,\mathrm{tail}}}{r_{t,\mathrm{tail}}}
    \right],
    \label{eq:grouped-kl}\\
    \mathcal L_{\mathrm{LM}}
    &=-\frac{1}{T}\sum_{t=1}^{T}\log r_{t,c_t},
    \qquad \mathcal L_{\mathrm{stage2}}
    =\mathcal L_{\mathrm{KL}}+\mathcal L_{\mathrm{LM}}.
    \label{eq:distillation-loss}
\end{align}
The two terms inside the brackets match the retained tokens and the total tail mass, respectively; individual tail tokens are not distinguished. The standard LM loss $\mathcal L_{\mathrm{LM}}$ increases the compressed-cache model's full-vocabulary probability of the teacher's greedy token $c_t=\arg\max_v p_{t,v}$. The two losses have equal weight and are averaged across sequences; cache size remains fixed.

\subsection{Teacher Rollouts for Document-Specific Supervision}
\label{sec:teacher-rollout}

A \emph{teacher rollout} is the frozen model's response to a generic prompt about document $D$. Let $C_D^{\mathrm{full}}$ be the cache obtained by prefilling $D$, and let $\{u_j\}_{j=1}^{J}$ be the fixed bank of $J=480$ text prompts (24 document requests paired with 20 response instructions; Appendix~\ref{app:source-prompts}). The prompts exclude downstream evaluation questions and answers.

\paragraph{Generating Rollouts.}
Each prompt starts from an independent fork of $C_D^{\mathrm{full}}$. With frozen model weights $\theta$, the teacher distribution and greedy continuation are
\begin{equation}
\begin{aligned}
    p_{j,t}(v)&=P_\theta(v\mid C_D^{\mathrm{full}},u_j,y_{j,<t}),\\
    y_{j,t}&=\arg\max_v p_{j,t}(v),\qquad 1\le t\le T_j\le24.
\end{aligned}
    \label{eq:teacher-rollout}
\end{equation}
Generation stops at an end-of-sequence token or after 24 tokens; prompts never share continuation history.

\paragraph{Using the Rollouts.}
Let $q_{\ell,h}^{(j,s)}\in\mathbb R^d$ be the teacher query at position $s$ while processing prompt $u_j$ and continuation $y_j$ in layer $\ell$ and query head $h$. Stage 1 collects
\begin{equation}
    \mathcal Q_{\ell,h}(D)=\{q_{\ell,h}^{(j,s)}:1\le j\le J,\ 1\le s\le |u_j|+T_j\}.
    \label{eq:rollout-queries}
\end{equation}
It matches full- and compressed-document attention outputs and log-normalizers on these queries using Equation~\ref{eq:attention-fitting-loss}, with preservers frozen. In stage 2, the compressed cache $C_D^{\mathrm{cmp}}$ predicts under the same teacher-generated prefix:
\begin{equation}
    r_{j,t}(v)=P_\theta(v\mid C_D^{\mathrm{cmp}},u_j,y_{j,<t}),\qquad
    \mathcal L_D=\frac1J\sum_{j=1}^{J}\mathcal L_{\mathrm{stage2}}\bigl((p_{j,t},r_{j,t})_{t=1}^{T_j}\bigr).
    \label{eq:rollout-distillation}
\end{equation}
Thus the student receives teacher tokens rather than its own predictions (teacher forcing). Equation~\ref{eq:distillation-loss} defines each rollout loss. Stored teacher top-20 logits and the vocabulary log-normalizer recover its top-20 probabilities and tail mass; both stages use the same fixed rollouts.

\begin{figure}[t]
    \centering
    \includegraphics[width=0.99\linewidth]{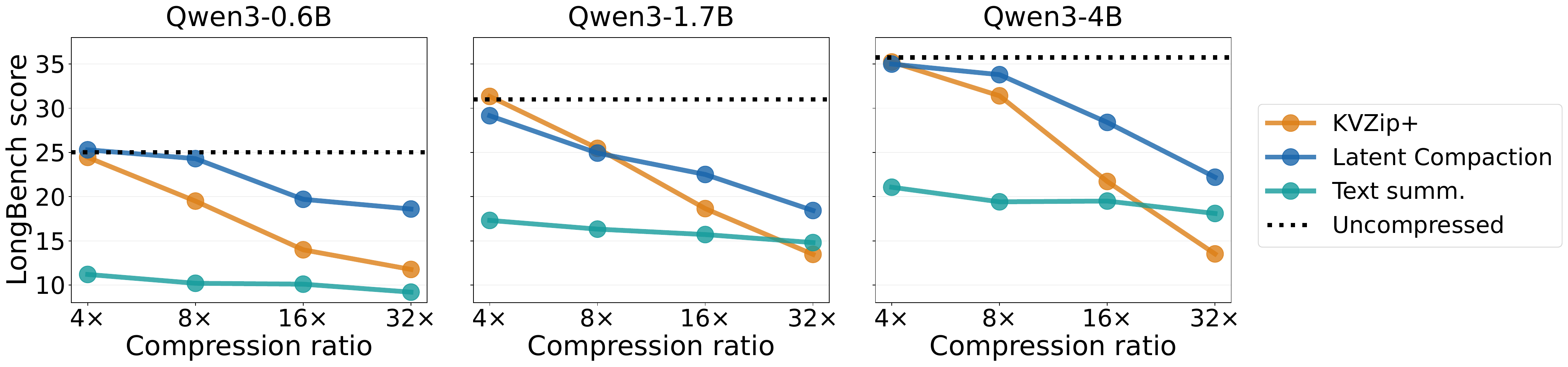}
    \caption{Comparison of compression methods on LongBench.}
    \label{fig:longbench-results}
\end{figure}

\begin{figure}[t]
    \includegraphics[width=0.99\linewidth]{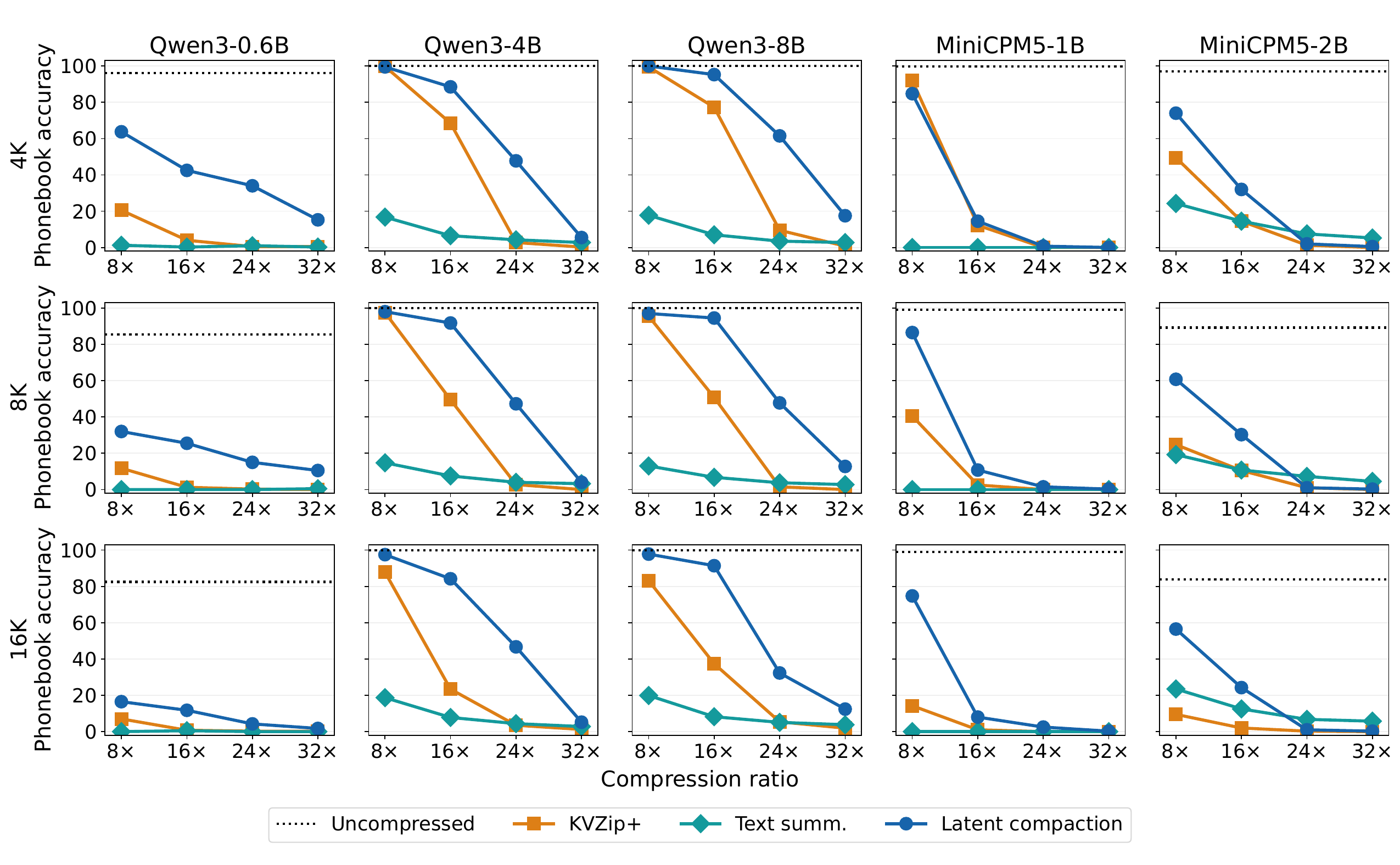}
    \caption{Phonebook Lookup results for additional Qwen3 model sizes and MiniCPM5 models. Each model uses its own document-specific cache optimization.}
    \label{fig:transfer-to-other-models}
\end{figure}

\begin{figure}[t]
    \centering
    \includegraphics[width=0.99\linewidth]{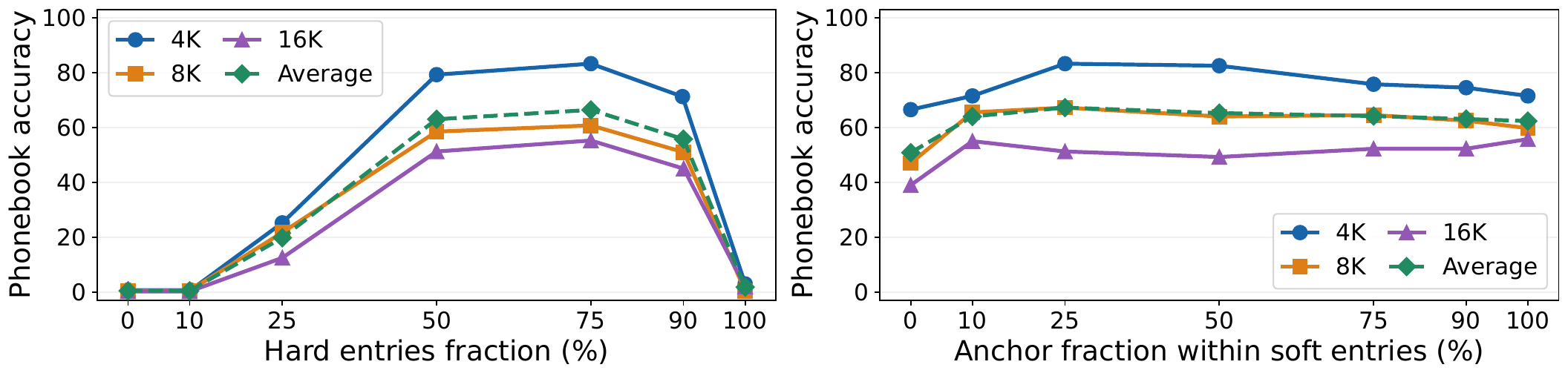}
    \caption{Phonebook Lookup~(4K, 8K, and 16K context lengths) accuracy of Latent Compaction at $16\times$ compression ratio with different fractions of slots allocated to preservers, and different anchor fractions within the combined anchor and diversifier budget.}
    \label{fig:slot-allocation}
\end{figure}

\section{Experiments}
\label{sec:experiments}

\subsection{Experimental setup}
\label{sec:experimental-setup}

\paragraph{Model and Training.}
Unless otherwise stated, we use Qwen3-1.7B with frozen model weights, BF16 model computation, and an FP32 KV cache. Each document receives 480 generic teacher prompts with continuations of up to 24 tokens. We use 1,000 attention-fitting (stage 1) updates per layer and 1,000 end-to-end refinement (stage 2) updates, with constant AdamW learning rates of $3\times10^{-3}$ and $10^{-4}$, respectively. We use one paired seed per document and fixed schedules without downstream-question-based checkpoint selection. Appendix~\ref{sec:experimental-details} gives implementation details.

\paragraph{Baselines and Budgets.}
We compare KVZip+, Attention Matching, SnapKV, and PyramidKV \citep{kim2025kvzip,jegou2026kvzap,zweiger2026attentionmatching,li2024snapkv,cai2024pyramidkv}, plus full-cache and text-summary references. Our Attention Matching baseline uses the repeat-prefill, highest-attention-key variant. Every method compresses the document before seeing the evaluation question. SnapKV and PyramidKV therefore use query-agnostic adaptations. Text summaries may fall below the target token budget; summaries that exceed it are truncated to that budget. For target context length $N_0$ and compression ratio $r$, the source budget is $B=N_0/r$ KV pairs per layer and KV head, counting all three entry types for Latent Compaction. Reported ratios use this target length; for an actual document length $N$, the KV-pair compression ratio is $N/B$. Question and generated-token caches are excluded. This is a logical KV-pair budget; physical memory also depends on precision, biases, and padding. 

\subsection{Evaluation benchmarks and metrics}
\label{sec:evaluation-benchmarks}

We evaluate on three benchmarks:
\begin{itemize}
    \item \textbf{RULER}~\citep{hsieh2024ruler}. This suite contains 13 synthetic long-context tasks. For each compression ratio, we evaluate 20 documents per task--length pair and report aggregate and per-task scores. Appendix~\ref{sec:ruler-results-detailed} reports the per-task results.
    \item \textbf{LongBench}~\citep{longbench}. We use this benchmark to evaluate long-context understanding on natural documents. We vary the compression ratio while keeping the evaluation documents fixed and report LongBench scores.
    \item \textbf{Phonebook Lookup}~\citep{repeat-after-me}. Each document maps names to phone numbers. Given a name, the model must retrieve the corresponding number. We report lookup accuracy.
\end{itemize}

Most ablations use Phonebook Lookup with 20 questions per document. Reusing each optimized document cache across these questions reduces evaluation cost. We use greedy decoding in all experiments to ensure reproducibility.

\begin{table}[t]
    \centering
    \small
    \caption{Initialization ablations on Phonebook Lookup with
    Qwen3-1.7B at $16\times$ compression.}
    \label{tab:phonebook-initialization}
    \begin{tabular}{l|cccc}
        \toprule
        Configuration & 4K & 8K & 16K & Avg. \\
        \midrule
        Latent Compaction (25\% anchors + 75\% diversifiers)      & \textbf{87.75} & \textbf{83.75} & \textbf{70.50} & \textbf{80.67} \\
        $\hookrightarrow$ Replace diversifiers with random document KVs & 29.25 & 20.00 & 12.25 & 20.50 \\
        $\hookrightarrow$ Replace diversifiers with Gaussian random vectors  & 27.50 & 17.50 & 14.25 & 19.75 \\
        $\hookrightarrow$ Replace diversifiers with first document KVs  & 65.50 & 42.50 & 22.50 & 43.50 \\
        $\hookrightarrow$ Replace anchors with KVZip+  & 69.25 & 58.75 & 43.50 & 57.17 \\
        \bottomrule
    \end{tabular}
\end{table}

\begin{table}[t]
    \centering
    \small
    \caption{Stage-1 ablation results on Phonebook Lookup with Qwen3-1.7B at $16\times$ compression.}
    \label{tab:ablation-results-stage1}
    \begin{tabular}{l|cccc}
        \toprule
        Method & 4K & 8K & 16K & Avg.\\
        \midrule
        Latent Compaction & \textbf{87.75} & \textbf{83.75} & \textbf{70.50} & \textbf{80.67} \\
        $\hookrightarrow$ Remove stage 1 & 13.75 & 13.25 & 14.75 & 13.92 \\
        $\hookrightarrow$ Remove frozen preserver KVs in stage 1 & 83.25 & 60.75 & 43.50 & 62.50 \\
        $\hookrightarrow$ Preserver KVs are trainable in stage 1 & 43.75 & 32.25 & 24.25 & 33.42 \\
        $\hookrightarrow$ Remove $\log Z$ MSE loss term & 42.50 & 38.00 & 29.00 & 36.50 \\
        \bottomrule
    \end{tabular}
\end{table}

\subsection{Main Results}
\label{sec:main-results}

\paragraph{RULER and Phonebook Lookup}

Figure~\ref{fig:main-results} compares methods across compression ratios at 16K context and across context lengths at $16\times$ and $32\times$ compression. Latent Compaction generally retains more accuracy than the compression baselines, with particularly large gains on Phonebook Lookup at $16\times$ compression. The advantage is not uniform: text summarization outperforms Latent Compaction on RULER at the most aggressive ratios, and all compressed methods approach zero Phonebook Lookup accuracy at $64\times$ compression and beyond.

\paragraph{LongBench}

Figure~\ref{fig:longbench-results} compares LongBench scores for Qwen3-0.6B, 1.7B, and 4B at $4\times$--$32\times$ compression. Latent Compaction outperforms text summarization throughout and KVZip+ at $16\times$ and $32\times$ for all three models. At $4\times$, its scores are close to KVZip+, but its advantage grows under stronger compression, particularly for the 0.6B and 4B models. Scores still decline as compression increases, leaving a gap to the uncompressed models. These results extend the retrieval findings to natural-document understanding, while showing the cost of aggressive compression.

\subsection{Transfer to Other Model Scales and Model Families}
\label{sec:transfer-to-other-model-scales}

We also apply the procedure to other model scales of the same family, 0.6B, 4B, and 8B, and to two models from another family, MiniCPM5~\citep{minicpm5}. For each model, cache optimization uses that model's own teacher outputs; this evaluates the applicability of the procedure across models rather than transfer of an optimized cache between models.
Figure~\ref{fig:transfer-to-other-models} compares Latent Compaction with KVZip+ and text summarization on Phonebook Lookup. The results show that its retrieval advantage extends to the additional model sizes and families.

\subsection{Ablations: Initialization}

\paragraph{Budget Allocation to Different Entries.}

Figure~\ref{fig:slot-allocation} varies the global budget fraction allocated to KVZip+-selected preservers (left) and the anchor fraction within the combined anchor and diversifier budget (right). Preserver allocation has the larger effect on accuracy; 25\% anchors gives the highest average among the fractions tested.

\paragraph{Initialization of Anchors and Diversifiers}

The default mixture of 25\% anchors and 75\% diversifiers outperforms every tested replacement initializer at each context length (Table~\ref{tab:phonebook-initialization}).

\subsection{Ablations: Training Stage 1}
\label{sec:ablations-stage-1}

Table~\ref{tab:ablation-results-stage1} shows that omitting stage 1 sharply reduces lookup accuracy at every context length. Allowing preserver keys and values to change during this stage also lowers accuracy substantially. Removing the log-normalizer loss causes a similarly large drop, supporting the use of both fixed preservers and normalization supervision during attention fitting.

\subsection{Ablations: Training Stage 2}
\label{sec:ablations-stage-2}

\begin{table}[t]
    \centering
    \small
    \caption{Stage-2 ablations on Phonebook Lookup with Qwen3-1.7B at $16\times$ compression. The default groups the teacher distribution into its top 20 tokens and the remaining mass.}
    \label{tab:ablation-results-stage2}
    \begin{tabular}{l|cccc}
        \toprule
        Method & 4K & 8K & 16K & Avg. \\
        \midrule
        Latent Compaction & \textbf{87.75} & 83.75 & \textbf{70.50} & \textbf{80.67} \\
        $\hookrightarrow$ Remove stage 2 & 81.00 & 68.25 & 62.75  & 70.67 \\
        $\hookrightarrow$ KL-divergence only & 83.00 & 83.50 & 65.25  & 77.25 \\
        $\hookrightarrow$ Remove tail-probability term in KL divergence & 82.00 & 81.75 & 61.00  & 75.25 \\
        $\hookrightarrow$ Full-vocabulary distillation & \textbf{87.75} & \textbf{84.25} & 70.00 & \textbf{80.67} \\
        \bottomrule
    \end{tabular}
\end{table}

Table~\ref{tab:ablation-results-stage2} shows that stage 2 improves accuracy by 6.75, 15.50, and 7.75 percentage points at 4K, 8K, and 16K, respectively. Adding language modeling loss $(\mathcal L_{\text{lm}})$ to grouped KL improves all three results, by 4.75, 0.25, and 5.25 points. Replacing the top-20-plus-tail targets with full-vocabulary targets leaves 4K accuracy unchanged, improves 8K accuracy by 0.50 points, and reduces 16K accuracy by 0.50 points. Thus, full-vocabulary distillation offers no consistent gain; the grouped objective retains similar accuracy with less teacher-target storage.

\paragraph{Training Duration}

\begin{figure}[!t]
    \centering
    \includegraphics[width=0.99\linewidth]{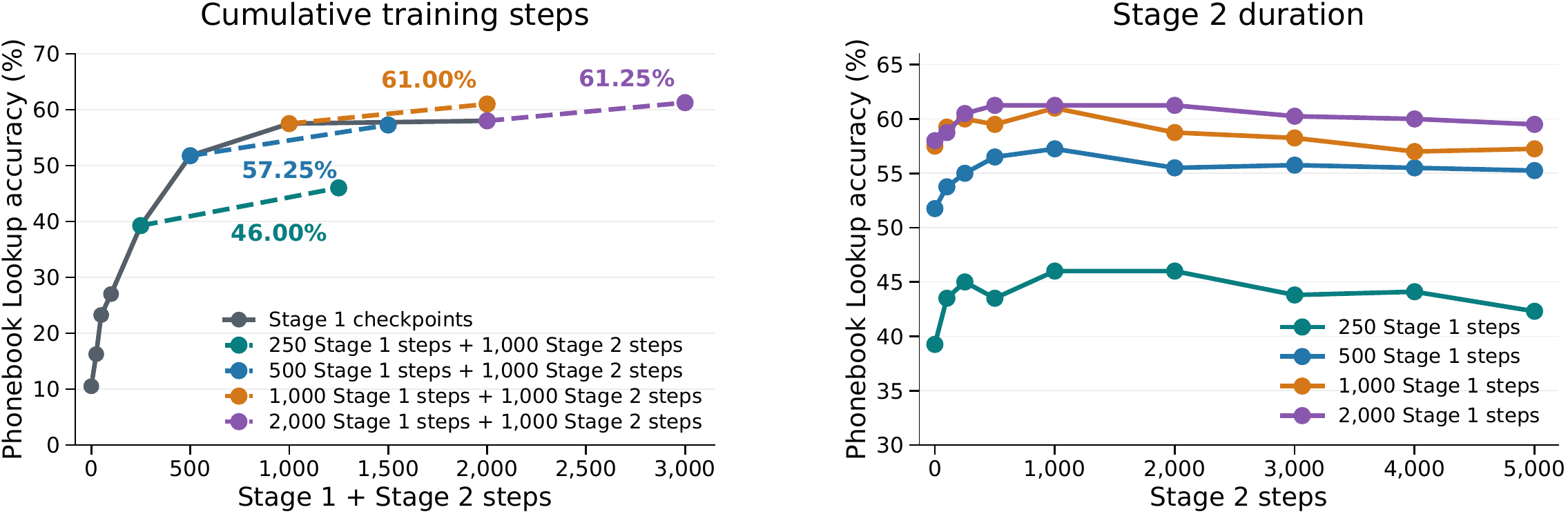}
    \caption{Phonebook Lookup accuracy of Latent Compaction of varying training durations. For clarity, dotted lines only show the accuracy at step 0 and step 1,000 of stage 2.}
    \label{fig:results-training-duration}
\end{figure}

Figure~\ref{fig:results-training-duration} shows how training duration affects accuracy. With stage 2 fixed at 1,000 updates, doubling stage 1 from 1,000 to 2,000 updates improves accuracy only from 61.00\% to 61.25\%. Longer stage-2 training is worse: across different stage-1 durations, accuracy peaks at 1,000 stage-2 updates and then declines slightly, showing signs of overfitting. Overall, this implies that using 1,000 updates for both stages offers a good balance of accuracy and preparation cost.

\begin{table}[!t]
    \centering
    \small
    \caption{Architectural ablations on Phonebook Lookup at $16\times$ compression.}
    \label{tab:model-ablations}
    \begin{tabular}{l|cccc}
    \toprule
    Method & 4K & 8K & 16K & Avg. \\
    \midrule
    Latent Compaction & \textbf{43.75} & \textbf{32.25} & \textbf{24.25} & \textbf{33.42} \\
    $\hookrightarrow$ Remove attention biases $(b_{\mathrm{cmp}})$ & 43.00 & 29.25 & 24.00 & 32.08 \\
    $\hookrightarrow$ Equal mix: replace Eq.~\ref{eq:autoregressive-attention} with $\tfrac12(A_{\mathrm{cmp}}+A_{\mathrm{new}})$ & 0.25 & 0.00 & 0.25 & 0.17 \\
    \bottomrule
    \end{tabular}
\end{table}

\subsection{Ablations: Model Architecture}

In Table~\ref{tab:model-ablations}, removing attention biases reduces accuracy by 0.75, 3.00, and 0.25 percentage points at 4K, 8K, and 16K. Replacing normalizer weighting~(Equation~\ref{eq:autoregressive-attention}) with an equal average at inference reduces accuracy to at most 0.25\%, despite retaining the stage-1 normalizer loss. Normalizer weighting recovers joint attention over document and subsequent entries.

\section{Conclusion}

Latent Compaction trains reusable document KV caches with frozen model weights through attention fitting and distillation. RULER, LongBench, and Phonebook Lookup show accuracy gains under compression; ablations highlight initialization and attention normalization. Per-document fitting remains costly and extreme compression hurts accuracy. Future work will reduce preparation cost and study broader tasks.

\subsection*{AI use statement}

We used generative AI to assist with manuscript editing and code implementation. We manually reviewed the implementation to check that it matches the intended method. We did not use generative AI to develop the research ideas, generate datasets, or originate mathematical claims. We take responsibility for the final content of this work, including all AI-assisted text, claims, and artifacts.

\FloatBarrier
\bibliography{iclr2027_conference}

@misc{longbench,
      title={LongBench: A Bilingual, Multitask Benchmark for Long Context Understanding}, 
      author={Yushi Bai and Xin Lv and Jiajie Zhang and Hongchang Lyu and Jiankai Tang and Zhidian Huang and Zhengxiao Du and Xiao Liu and Aohan Zeng and Lei Hou and Yuxiao Dong and Jie Tang and Juanzi Li},
      year={2024},
      eprint={2308.14508},
      archivePrefix={arXiv},
      primaryClass={cs.CL},
      url={https://arxiv.org/abs/2308.14508}, 
}

@misc{zhang2023h2o,
  title = {{{H$_2$O}: Heavy-Hitter Oracle for Efficient Generative Inference of Large Language Models}},
  author = {Zhang, Zhenyu and Sheng, Ying and Zhou, Tianyi and Chen, Tianlong and Zheng, Lianmin and Cai, Ruisi and Song, Zhao and Tian, Yuandong and R{\'e}, Christopher and Barrett, Clark and Wang, Zhangyang and Chen, Beidi},
  year = {2023},
  eprint = {2306.14048},
  archivePrefix = {arXiv},
  primaryClass = {cs.LG},
  url = {https://arxiv.org/abs/2306.14048}
}

@misc{li2024snapkv,
  title = {{{SnapKV}: {LLM} Knows What You are Looking for Before Generation}},
  author = {Li, Yuhong and Huang, Yingbing and Yang, Bowen and Venkitesh, Bharat and Locatelli, Acyr and Ye, Hanchen and Cai, Tianle and Lewis, Patrick and Chen, Deming},
  year = {2024},
  eprint = {2404.14469},
  archivePrefix = {arXiv},
  primaryClass = {cs.CL},
  url = {https://arxiv.org/abs/2404.14469}
}

@misc{cai2024pyramidkv,
  title = {{{PyramidKV}: Dynamic {KV} Cache Compression based on Pyramidal Information Funneling}},
  author = {Cai, Zefan and Zhang, Yichi and Gao, Bofei and Liu, Yuliang and Li, Yucheng and Liu, Tianyu and Lu, Keming and Xiong, Wayne and Dong, Yue and Hu, Junjie and Xiao, Wen},
  year = {2025},
  eprint = {2406.02069},
  archivePrefix = {arXiv},
  primaryClass = {cs.CL},
  url = {https://arxiv.org/abs/2406.02069}
}

@misc{feng2024adakv,
  title = {{{Ada-KV}: Optimizing {KV} Cache Eviction by Adaptive Budget Allocation for Efficient {LLM} Inference}},
  author = {Feng, Yuan and Lv, Junlin and Cao, Yukun and Xie, Xike and Zhou, S. Kevin},
  year = {2025},
  eprint = {2407.11550},
  archivePrefix = {arXiv},
  primaryClass = {cs.CL},
  url = {https://arxiv.org/abs/2407.11550}
}

@misc{kim2025kvzip,
  title = {{{KVzip}: Query-Agnostic {KV} Cache Compression with Context Reconstruction}},
  author = {Kim, Jang-Hyun and Kim, Jinuk and Kwon, Sangwoo and Lee, Jae W. and Yun, Sangdoo and Song, Hyun Oh},
  year = {2025},
  eprint = {2505.23416},
  archivePrefix = {arXiv},
  primaryClass = {cs.DB},
  url = {https://arxiv.org/abs/2505.23416}
}

@misc{li2021prefixtuning,
  title = {{{Prefix-Tuning}: Optimizing Continuous Prompts for Generation}},
  author = {Li, Xiang Lisa and Liang, Percy},
  year = {2021},
  eprint = {2101.00190},
  archivePrefix = {arXiv},
  primaryClass = {cs.CL},
  url = {https://arxiv.org/abs/2101.00190}
}

@misc{chevalier2023autocompressors,
  title = {{Adapting Language Models to Compress Contexts}},
  author = {Chevalier, Alexis and Wettig, Alexander and Ajith, Anirudh and Chen, Danqi},
  year = {2023},
  eprint = {2305.14788},
  archivePrefix = {arXiv},
  primaryClass = {cs.CL},
  url = {https://arxiv.org/abs/2305.14788}
}

@misc{ge2023icae,
  title = {{In-context Autoencoder for Context Compression in a Large Language Model}},
  author = {Ge, Tao and Hu, Jing and Wang, Lei and Wang, Xun and Chen, Si-Qing and Wei, Furu},
  year = {2024},
  eprint = {2307.06945},
  archivePrefix = {arXiv},
  primaryClass = {cs.CL},
  url = {https://arxiv.org/abs/2307.06945}
}

@misc{eyuboglu2025cartridges,
  title = {{Cartridges: Lightweight and general-purpose long context representations via self-study}},
  author = {Eyuboglu, Sabri and Ehrlich, Ryan and Arora, Simran and Guha, Neel and Zinsley, Dylan and Liu, Emily and Tennien, Will and Rudra, Atri and Zou, James and Mirhoseini, Azalia and Re, Christopher},
  year = {2025},
  eprint = {2506.06266},
  archivePrefix = {arXiv},
  primaryClass = {cs.CL},
  url = {https://arxiv.org/abs/2506.06266}
}

@misc{zweiger2026attentionmatching,
  title = {{Fast {KV} Compaction via Attention Matching}},
  author = {Zweiger, Adam and Fu, Xinghong and Guo, Han and Kim, Yoon},
  year = {2026},
  eprint = {2602.16284},
  archivePrefix = {arXiv},
  primaryClass = {cs.LG},
  url = {https://arxiv.org/abs/2602.16284}
}

@article{baek2026restorekv,
  title = {{{RestoreKV}: Recovering Full-Cache Behavior Under Aggressive Query-Agnostic {KV} Cache Eviction}},
  author = {Baek, Changwoo and Shin, Seungjun and Kong, Kyeongbo},
  journal = {arXiv preprint arXiv:2608.01247},
  year = {2026}
}

@misc{liu2024kivi,
  title = {{{KIVI}: A Tuning-Free Asymmetric 2bit Quantization for {KV} Cache}},
  author = {Liu, Zirui and Yuan, Jiayi and Jin, Hongye and Zhong, Shaochen and Xu, Zhaozhuo and Braverman, Vladimir and Chen, Beidi and Hu, Xia},
  year = {2024},
  eprint = {2402.02750},
  archivePrefix = {arXiv},
  primaryClass = {cs.CL},
  doi = {https://doi.org/10.13140/RG.2.2.28167.37282},
  url = {https://arxiv.org/abs/2402.02750}
}

@article{gonzalez1985clustering,
  author = {Gonzalez, Teofilo F.},
  title = {Clustering to Minimize the Maximum Intercluster Distance},
  journal = {Theoretical Computer Science},
  volume = {38},
  pages = {293--306},
  year = {1985},
  doi = {10.1016/0304-3975(85)90224-5},
  url = {https://doi.org/10.1016/0304-3975(85)90224-5}
}

@misc{sun2020testtimetraining,
  title = {{Test-Time Training with Self-Supervision for Generalization under Distribution Shifts}},
  author = {Yu Sun and Xiaolong Wang and Zhuang Liu and John Miller and Alexei A. Efros and Moritz Hardt},
  year = {2020},
  eprint = {1909.13231},
  archivePrefix = {arXiv},
  primaryClass = {cs.LG},
  url = {https://arxiv.org/abs/1909.13231}
}

@inproceedings{hardt2024tttnn,
  title = {{Test-Time Training on Nearest Neighbors for Large Language Models}},
  author = {Hardt, Moritz and Sun, Yu},
  booktitle = {International Conference on Learning Representations},
  year = {2024},
  url = {https://arxiv.org/abs/2305.18466}
}

@misc{sun2025tttlayers,
  title = {{Learning to (Learn at Test Time): RNNs with Expressive Hidden States}},
  author = {Yu Sun and Xinhao Li and Karan Dalal and Jiarui Xu and Arjun Vikram and Genghan Zhang and Yann Dubois and Xinlei Chen and Xiaolong Wang and Sanmi Koyejo and Tatsunori Hashimoto and Carlos Guestrin},
  year = {2025},
  eprint = {2407.04620},
  archivePrefix = {arXiv},
  primaryClass = {cs.LG},
  url = {https://arxiv.org/abs/2407.04620}
}

@misc{tandon2025ttte2e,
  title = {{End-to-End Test-Time Training for Long Context}},
  author = {Tandon, Arnuv and Dalal, Karan and Li, Xinhao and Koceja, Daniel and R{\o}d, Marcel and Buchanan, Sam and Wang, Xiaolong and Leskovec, Jure and Koyejo, Sanmi and Hashimoto, Tatsunori and Guestrin, Carlos and McCaleb, Jed and Choi, Yejin and Sun, Yu},
  year = {2025},
  eprint = {2512.23675},
  archivePrefix = {arXiv},
  primaryClass = {cs.LG},
  url = {https://arxiv.org/abs/2512.23675}
}

@misc{hsieh2024ruler,
  title = {{{RULER}: What's the Real Context Size of Your Long-Context Language Models?}},
  author = {Hsieh, Cheng-Ping and Sun, Simeng and Kriman, Samuel and Acharya, Shantanu and Rekesh, Dima and Jia, Fei and Zhang, Yang and Ginsburg, Boris},
  year = {2024},
  eprint = {2404.06654},
  archivePrefix = {arXiv},
  primaryClass = {cs.CL},
  url = {https://arxiv.org/abs/2404.06654}
}

@misc{repeat-after-me,
      title={Repeat After Me: Transformers are Better than State Space Models at Copying}, 
      author={Samy Jelassi and David Brandfonbrener and Sham M. Kakade and Eran Malach},
      year={2024},
      eprint={2402.01032},
      archivePrefix={arXiv},
      primaryClass={cs.LG},
      url={https://arxiv.org/abs/2402.01032}, 
}

@misc{minicpm5,
  title={{MiniCPM5}: Small Language Models},
  author={{OpenBMB}},
  year={2026},
  howpublished={\url{https://github.com/OpenBMB/MiniCPM}},
  note={Official model repository, accessed September 26, 2026}
}

@inproceedings{xiao2024efficientstreaminglanguagemodels,
  title={Efficient Streaming Language Models with Attention Sinks},
  author={Xiao, Guangxuan and Tian, Yuandong and Chen, Beidi and Han, Song and Lewis, Mike},
  booktitle={International Conference on Learning Representations},
  year={2024},
  url={https://openreview.net/forum?id=NG7sS51zVF}
}

@misc{anthropic2026compaction,
  author={{Anthropic}},
  title={Compaction at a Token Threshold},
  year={2026},
  howpublished={Claude Platform documentation},
  url={https://platform.claude.com/docs/en/build-with-claude/compaction-threshold},
  note={Accessed September 26, 2026}
}

@misc{openai2026compaction,
  author={{OpenAI}},
  title={Compaction},
  year={2026},
  howpublished={OpenAI API documentation},
  url={https://developers.openai.com/api/docs/guides/compaction},
  note={Accessed September 26, 2026}
}

@misc{anthropic2025contextengineering,
  author={{Anthropic}},
  title={Effective Context Engineering for {AI} Agents},
  year={2025},
  url={https://www.anthropic.com/engineering/effective-context-engineering-for-ai-agents}
}

@misc{jegou2026kvzap,
  title={{KVzap}: Fast, Adaptive, and Faithful {KV} Cache Pruning},
  author={J{\'e}gou, Simon and Jeblick, Maximilian},
  year={2026},
  eprint={2601.07891},
  archivePrefix={arXiv},
  primaryClass={cs.LG},
  url={https://arxiv.org/abs/2601.07891}
}
\bibliographystyle{iclr2027_conference}

\appendix

\section{Additional Experimental Details}
\label{sec:experimental-details}

\subsection{Model Configuration}
All main experiments use the same Qwen3-1.7B reader checkpoint and tokenizer. The model has 28 Transformer layers, hidden dimension 2,048, 16 query heads, eight KV heads, and head dimension 128. Model weights remain frozen. We use BF16 inference with the original rotary positional configuration, without extending the context window or modifying the architecture. Inputs use raw text, and checkpoint and tokenizer hashes are verified across hosts.

\begin{center}
\begin{minipage}{\linewidth}
\centering
\small
\captionof{table}{Hugging Face checkpoints used in the experiments. Each model uses its corresponding tokenizer.}
\label{tab:model-checkpoints}
\begin{tabular}{@{}ll@{}}
\toprule
Model & Checkpoint URL \\
\midrule
Qwen3-0.6B & \url{https://huggingface.co/Qwen/Qwen3-0.6B} \\
Qwen3-1.7B & \url{https://huggingface.co/Qwen/Qwen3-1.7B} \\
Qwen3-4B & \url{https://huggingface.co/Qwen/Qwen3-4B} \\
Qwen3-8B & \url{https://huggingface.co/Qwen/Qwen3-8B} \\
MiniCPM5-1B & \url{https://huggingface.co/openbmb/MiniCPM5-1B} \\
MiniCPM5-2B & \url{https://huggingface.co/openbmb/MiniCPM5-2B} \\
\bottomrule
\end{tabular}
\end{minipage}
\end{center}

\subsection{Document-Specific Training}

For each document, the full-cache model generates up to 24 tokens for each of 480 fixed generic prompts. We reuse these teacher rollouts in both stages. Model weights are frozen; the document-cache parameters are optimized in FP32, while model computation uses BF16. Table~\ref{tab:training-implementation} summarizes the reported configuration, which uses 1,000 updates per stage by default.

\begin{center}
\begin{minipage}{\linewidth}
\centering
\small
\captionof{table}{Document-specific optimization settings. A stage-2 batch consists of four independently processed rollouts whose gradients are accumulated before one update.}
\label{tab:training-implementation}
\begin{tabular}{@{}lcc@{}}
\toprule
Setting & Stage 1: attention fitting & Stage 2: end-to-end refinement \\
\midrule
Updates & 1,000 per layer & 1,000 jointly across layers \\
Examples per update & up to 128 query positions & 4 teacher rollouts \\
AdamW learning rate & $3\times10^{-3}$ & $10^{-4}$ \\
Learning-rate schedule & constant; no warmup & constant; no warmup \\
Weight decay & 0 & 0 \\
Gradient norm clip & 1 & 1 \\
\bottomrule
\end{tabular}
\end{minipage}
\end{center}

AdamW uses its default momentum parameters $\beta_1=0.9$, $\beta_2=0.999$, and $\epsilon=10^{-8}$. There is no learning-rate decay or gradient accumulation across optimizer updates. Stage 1 trains anchor and diversifier keys, values, and biases, while preserver keys, values, and biases remain frozen. Stage 2 trains the entire compacted cache, including preserver keys and values and all attention biases. Both stages use fixed update counts; no checkpoint is selected using downstream questions. The same document seed is shared across paired conditions.

\paragraph{Example Construction and Batching}
\begin{itemize}
    \item \textbf{Document and teacher prompts.} We tokenize each document as raw text, without a chat template or added special tokens, and prefill it once. We do not concatenate different documents or teacher prompts into a single sequence. Each of the 480 prompts starts from an independent fork of that document's full cache; generation stops at EOS or 24 tokens. The fitting code does not truncate document or prompt tokens, so inputs must fit the model's supported context length.
    \item \textbf{Stage 1.} Query vectors captured from all prompt--continuation rollouts are concatenated along the query-position axis for each layer. If this pool exceeds 4,096 positions, we retain 4,096 evenly spaced positions. Each update samples up to 128 query positions with replacement, shared across the layer's heads. Full-cache attention targets are computed in chunks of 16 query positions to limit temporary memory.
    \item \textbf{Stage 2.} Each update samples four rollouts with replacement. For each rollout, the input is its prompt concatenated with the preceding teacher-generated tokens; the document is supplied through the compacted cache. The model processes each rollout separately with a fresh continuation cache and a sequence batch size of one. Its loss is divided by four, and gradients from all four rollouts are accumulated before clipping and one AdamW step. Variable-length rollouts are neither padded together nor packed into a shared attention sequence. Distillation uses the teacher's top 20 probabilities and aggregate tail mass, plus next-token cross-entropy (Section~\ref{sec:training-stage-2}).
\end{itemize}

\paragraph{Attention Implementation}
The uncompressed teacher follows the model's native Hugging Face attention path. For the compacted cache, the adapter computes scaled query--key scores explicitly with PyTorch tensor operations, adds learned per-entry biases, masks padded preserver slots, and applies softmax to obtain the document attention output. It also computes the document log-normalizer with \texttt{logsumexp}. A separate causal PyTorch attention calculation covers prompt and generated-token entries; their output and log-normalizer are combined with the document terms using the shared softmax normalizer. Attention scores and normalizers are evaluated in FP32. The stage-1 full-cache targets likewise use explicit matrix multiplication, softmax, and \texttt{logsumexp}. This compacted-cache path does not call FlashAttention or PyTorch's fused scaled-dot-product attention: it needs the log-normalizer as well as the attention output, plus per-entry biases and a ragged-slot mask. A custom fused kernel that exposes these quantities could implement the same mathematics; FlashAttention is not inherently incompatible with the method.

\paragraph{Normalizer Computation and Storage}
The Python-style pseudocode below shows the per-layer, per-query-head computation behind Equation~\ref{eq:autoregressive-attention}. The compressed branch includes learned biases and excludes padded slots; the subsequent-token branch uses a causal visibility mask and no learned bias. Both branches compute attention outputs and log-normalizers in FP32.

\noindent\textbf{Python pseudocode: normalizer-aware cache attention}
\begin{lstlisting}[style=python]
import torch


def branch_attention(q, keys, values, bias, valid, scale):
    scores = scale * (q.float() @ keys.float().T)
    if bias is not None:
        scores = scores + bias.float()
    scores = scores.masked_fill(
        ~valid, torch.finfo(scores.dtype).min
    )
    output = torch.softmax(scores, dim=-1) @ values.float()
    log_z = torch.logsumexp(scores, dim=-1)
    return output, log_z


def combine_attention(q, doc_k, doc_v, doc_bias, doc_valid,
                      new_k, new_v, new_visible, scale):
    # q has RoPE; GQA keys/values are expanded to query heads.
    doc_out, doc_log_z = branch_attention(
        q, doc_k, doc_v, doc_bias, doc_valid, scale
    )
    new_out, new_log_z = branch_attention(
        q, new_k, new_v, None, new_visible, scale
    )
    total_log_z = torch.logaddexp(doc_log_z, new_log_z)
    doc_weight = torch.exp(doc_log_z - total_log_z)
    new_weight = torch.exp(new_log_z - total_log_z)
    return (doc_weight * doc_out + new_weight * new_out).to(q.dtype)
\end{lstlisting}

During stage 1, the full-cache target is $\log Z_{\mathrm{full}}=\operatorname{logsumexp}(s qK_{\mathrm{full}}^{\top})$; the compressed branch's \texttt{doc\_log\_z} is fitted to this target using Equation~\ref{eq:attention-fitting-loss}. The teacher mean and standard deviation are fixed per query head, with the latter floored at $10^{-4}$. Normalizers are recomputed for each query rather than saved in the cache artifact; keys, values, biases, and masks are saved. These attention normalizers differ from the vocabulary log-normalizer saved with teacher logits for stage-2 distillation.

The reference configuration assigns one quarter of the logical KV-pair budget to anchors and diversifiers, and the remainder to globally selected preservers. One quarter of the anchor--diversifier slots are anchors; the rest use farthest-point initialization. We train exactly the number of slots deployed. Teacher artifacts are reused across configurations for the same document. Evaluation accounts for BF16 deployment precision.

\subsection{Baselines and Budgets}
We compare with KVZip+, Attention Matching, SnapKV, and PyramidKV \citep{kim2025kvzip,jegou2026kvzap,zweiger2026attentionmatching,li2024snapkv,cai2024pyramidkv}, together with an uncompressed reference. Every compression method receives the document before its evaluation question is revealed. Attention Matching uses repeat-prefill reference queries, RMS attention-based key selection, two bounded nonnegative least-squares iterations with bounds $[e^{-3},e^3]$, and least-squares value fitting. This is the tested highest-attention-key variant, without on-policy query recollection or OMP. SnapKV and PyramidKV use a 32-token document observation window, mean pooling of width five, and shared GQA entries. PyramidKV uses $\beta=20$ and an allocation rounded to preserve the total budget exactly. Their results describe query-agnostic adaptations.

For target context length $N_0$ and compression ratio $r$, we set the average source budget to $B=N_0/r$ KV pairs per layer and KV head. Retained and latent entries both count toward this budget. We separately distinguish logical pair counts from physical memory, which also depends on log-weight metadata, padding, and precision. Query and continuation caches are outside the source budget.

\section{Training Loss Curves}
\label{app:training-loss-curves}

\begin{figure}[t]
    \centering
    \includegraphics[width=0.99\linewidth]{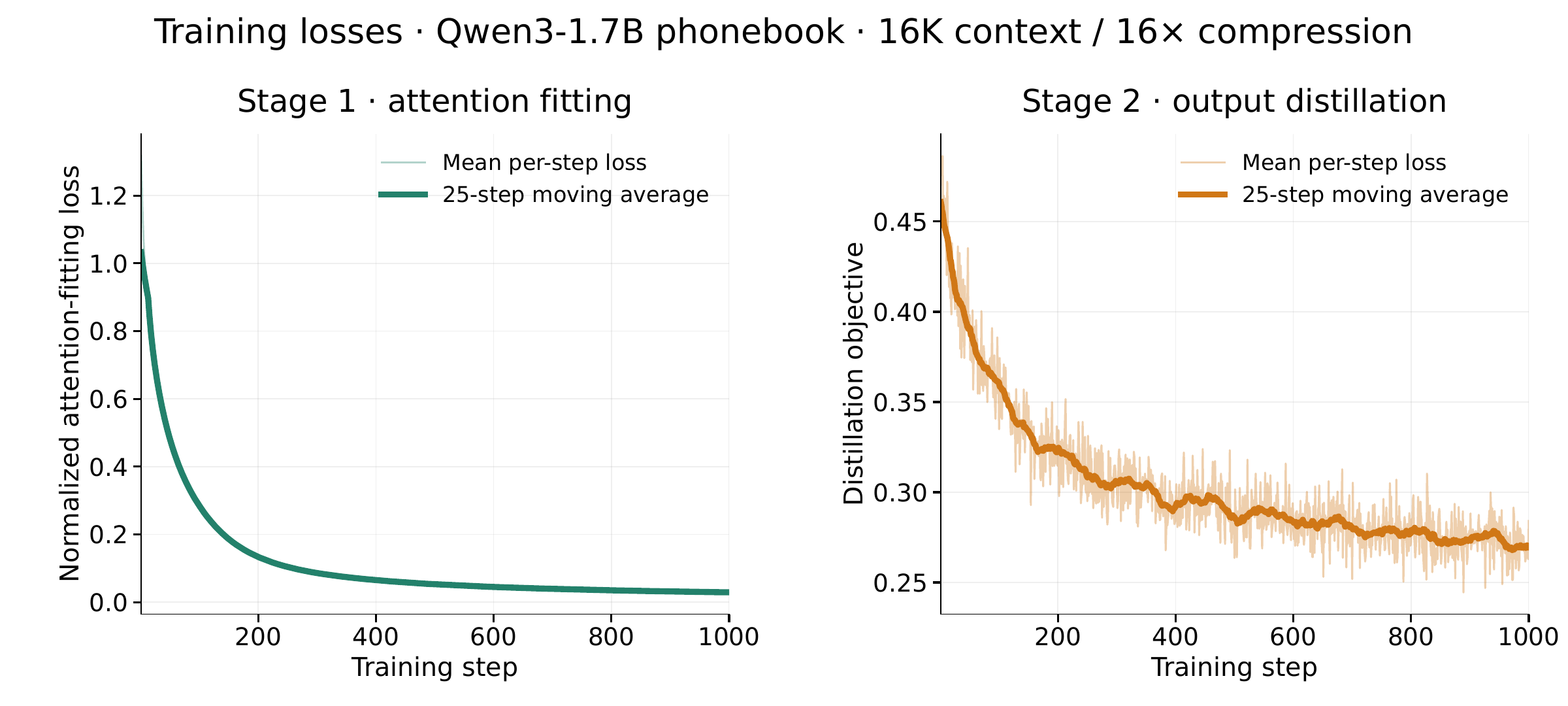}
    \caption{Qwen3-1.7B training losses on 16K Phonebook Lookup at $16\times$ compression. Stage 1 shows the normalized attention-fitting objective (Equation~\ref{eq:attention-fitting-loss}); stage 2 shows grouped KL and next-token cross-entropy (Equation~\ref{eq:distillation-loss}). Each stage runs for 1,000 updates. Thin lines show per-step means; thick lines show 25-step moving averages.}
    \label{fig:phonebook-training-losses}
\end{figure}

Figure~\ref{fig:phonebook-training-losses} shows the training objectives for a Qwen3-1.7B Phonebook Lookup run at 16K context and $16\times$ compression. The normalized attention-fitting loss in stage 1 decreases rapidly at first and continues to decline over 1,000 updates. The stage-2 distillation objective also decreases over its 1,000 updates, with greater per-step variation and a more gradual decline later in training. The two panels show different objectives, so their loss values are not directly comparable.

\FloatBarrier

\begin{figure}[t]
\centering
\setlength{\fboxsep}{7pt}
\setlength{\fboxrule}{0.5pt}
\fbox{\begin{minipage}{\dimexpr\linewidth-2\fboxsep-2\fboxrule\relax}
{\footnotesize\bfseries Document}\par\vspace{2pt}
{\small\ttfamily
\begin{tabular}{@{}l@{}}
The phone number of \textless NAME\_1\textgreater\ is \textless NUMBER\_1\textgreater.\\
The phone number of \textless NAME\_2\textgreater\ is \textless NUMBER\_2\textgreater.\\
...\\
The phone number of \textless NAME\_N\textgreater\ is \textless NUMBER\_N\textgreater.
\end{tabular}\par}
\vspace{5pt}
{\footnotesize\bfseries Question}\par\vspace{2pt}
{\small\ttfamily Question: What is the phone number of \textless NAME\_i\textgreater?\par}
\vspace{5pt}
{\footnotesize\bfseries Response prefix}\par\vspace{2pt}
{\small\ttfamily Answer:\par}
\end{minipage}}
\caption{Phonebook Lookup prompt template. Angle-bracketed fields are
placeholders, and the ellipsis denotes additional entries. The model
continues after \texttt{Answer:}. Bold section labels are explanatory and
are not part of the plain-text prompt.}
\label{fig:phonebook-prompt}
\end{figure}

\section{Prompt Template for Phonebook Lookup}

Figure~\ref{fig:phonebook-prompt} shows the prompt template used for Phonebook Lookup.

\paragraph{Phonebook Lookup Task}
Each document contains synthetic name--number pairs, with unique
10-character uppercase names and 10-digit phone numbers. We sample
20 distinct entries per document and evaluate each lookup independently.
The document is cached once and reused across questions; evaluation
questions are excluded from cache construction and optimization.

\section{Generic Prompts for Teacher Rollouts}
\label{app:source-prompts}

For each document, we use a fixed bank of 480 generic prompts to collect
training signals from the full-cache teacher. The bank is the Cartesian
product of 24 document-oriented requests and 20 response instructions. Each
prompt has the form
\begin{quote}
\ttfamily
<document request> <response instruction>\\
Answer:
\end{quote}
For example, one combination is:
\begin{quote}
\ttfamily
Summarize the document's main subject and structure. Answer briefly.\\
Answer:
\end{quote}
The requests cover broad properties such as content, organization, and
writing style. The response instructions vary the desired length and
specificity. The same prompt bank is used for every document; its text is
fixed in advance and contains neither benchmark questions nor their answers.

We first prefill the document into the frozen teacher's full KV cache. For
each prompt, we fork that cache and greedily generate up to 24 continuation
tokens. The resulting attention queries provide supervision for attention
fitting, while teacher next-token distributions along the generated
continuations provide targets for distillation. Evaluation questions are
held out until the compacted cache has been constructed and optimized.
Thus, the 480 prompts are alternative ways to probe a \emph{single}
document, rather than 480 additional training documents.

\section{RULER Results By Subtask}
\label{sec:ruler-results-detailed}

\begin{figure*}[t]
  \centering
  \includegraphics[width=\textwidth]{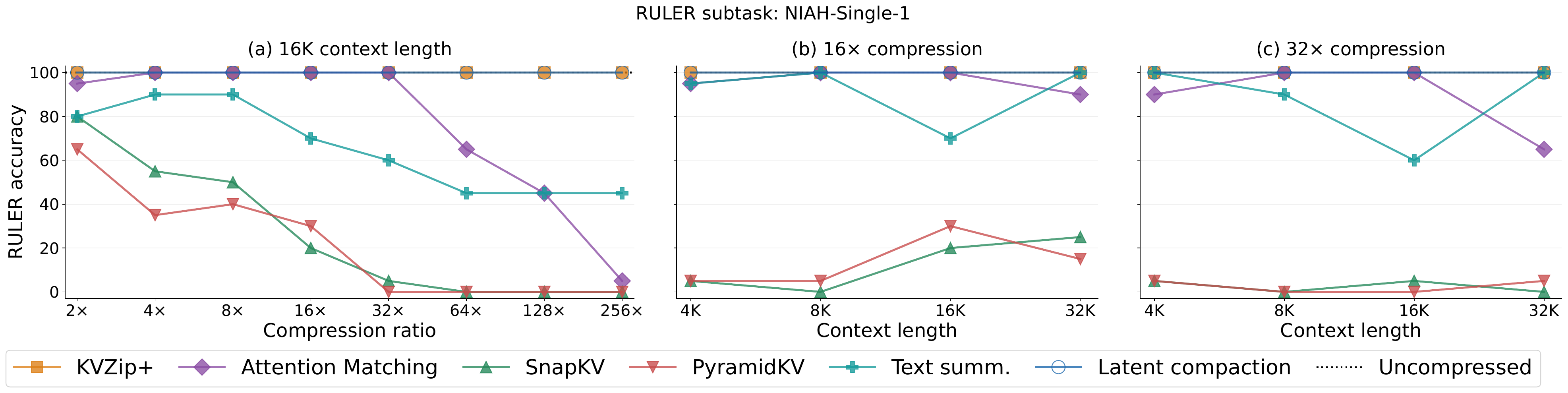}
  \caption{RULER NIAH-Single-1: accuracy versus compression ratio at 16K context (left), and versus context length at $16\times$ (middle) and $32\times$ compression (right). }
  \label{fig:ruler-niah-single-1}
\end{figure*}

\begin{figure*}[t]
  \centering
  \includegraphics[width=\textwidth]{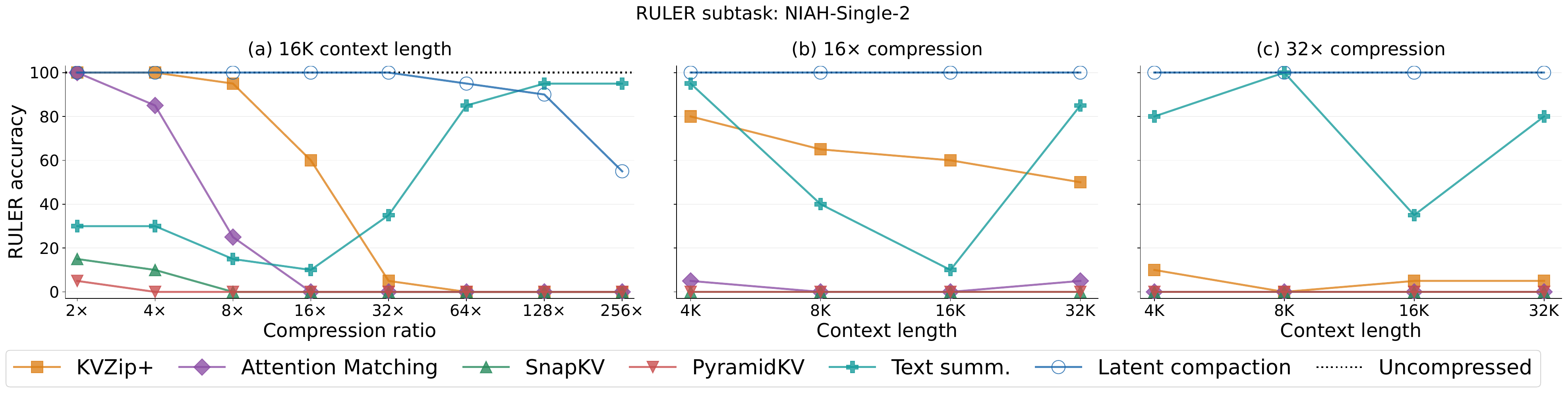}
  \caption{RULER NIAH-Single-2: accuracy versus compression ratio at 16K context (left), and versus context length at $16\times$ (middle) and $32\times$ compression (right). }
  \label{fig:ruler-niah-single-2}
\end{figure*}

\begin{figure*}[t]
  \centering
  \includegraphics[width=\textwidth]{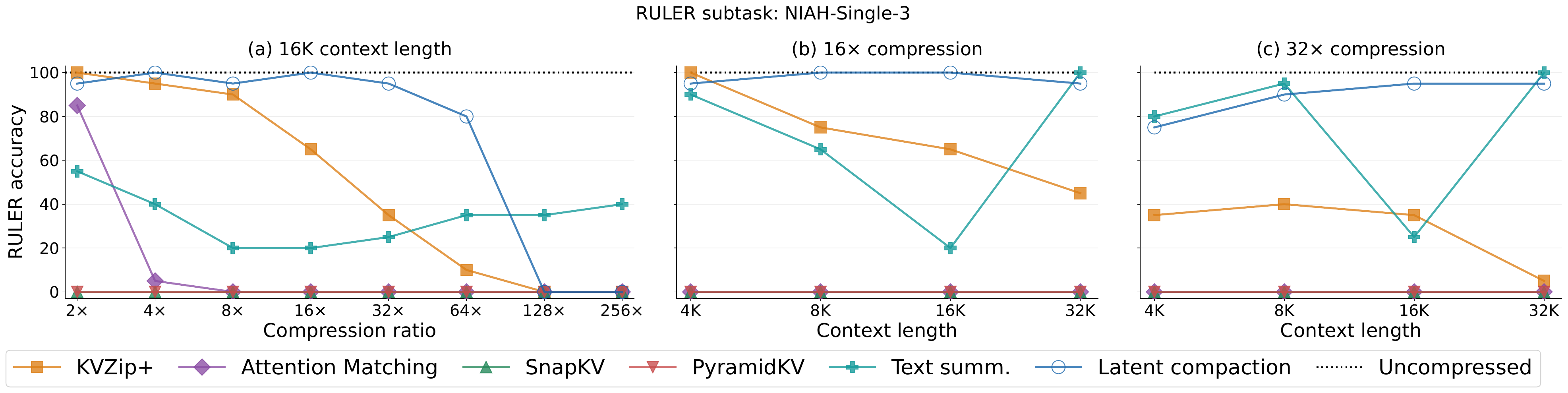}
  \caption{RULER NIAH-Single-3: accuracy versus compression ratio at 16K context (left), and versus context length at $16\times$ (middle) and $32\times$ compression (right). }
  \label{fig:ruler-niah-single-3}
\end{figure*}

\begin{figure*}[t]
  \centering
  \includegraphics[width=\textwidth]{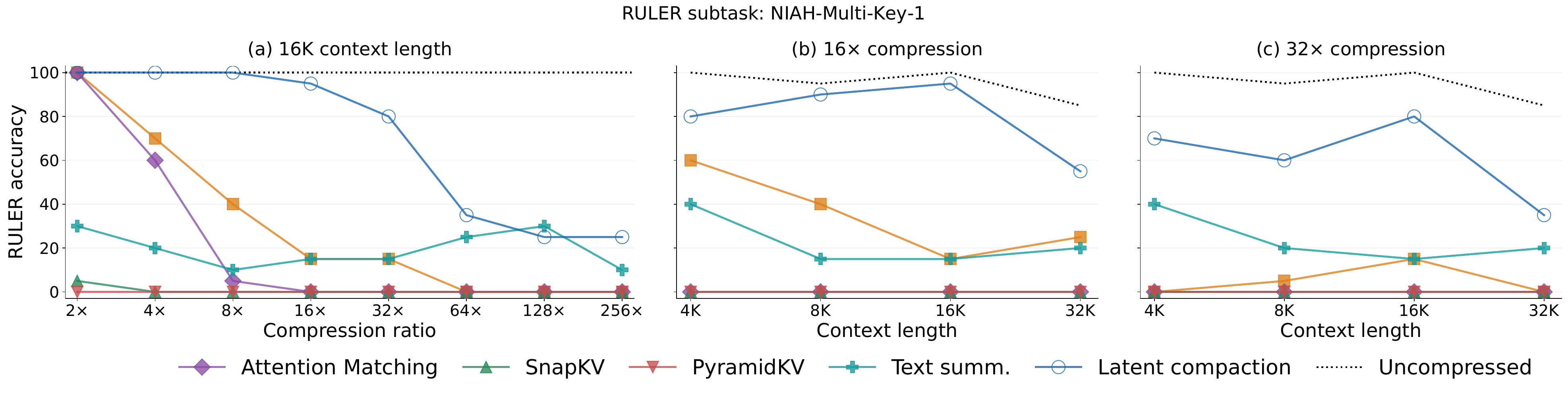}
  \caption{RULER NIAH-Multi-Key-1: accuracy versus compression ratio at 16K context (left), and versus context length at $16\times$ (middle) and $32\times$ compression (right). }
  \label{fig:ruler-niah-multikey-1}
\end{figure*}

\begin{figure*}[t]
  \centering
  \includegraphics[width=\textwidth]{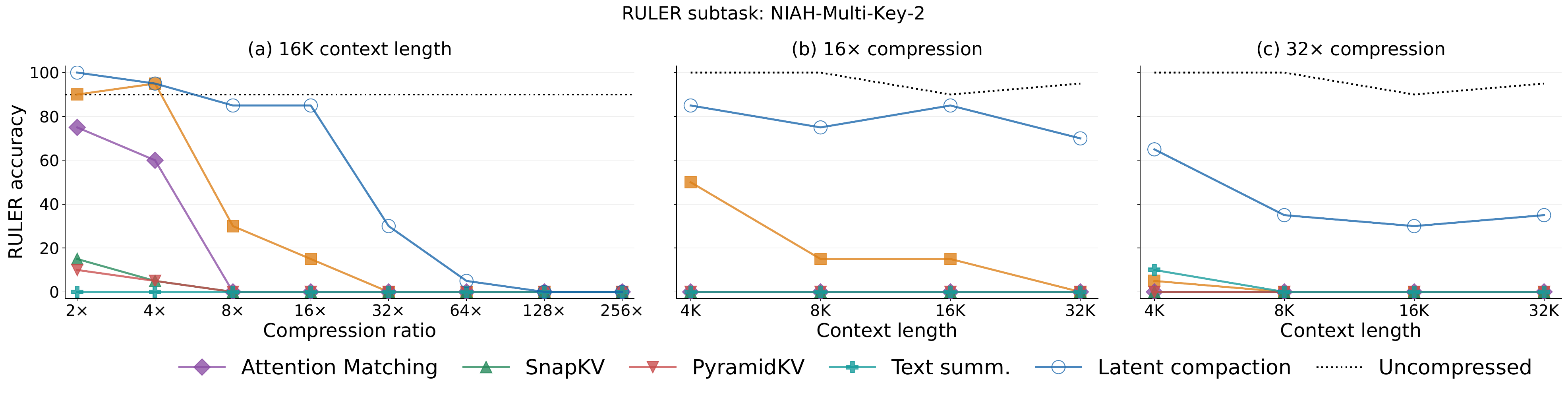}
  \caption{RULER NIAH-Multi-Key-2: accuracy versus compression ratio at 16K context (left), and versus context length at $16\times$ (middle) and $32\times$ compression (right). }
  \label{fig:ruler-niah-multikey-2}
\end{figure*}

\begin{figure*}[t]
  \centering
  \includegraphics[width=\textwidth]{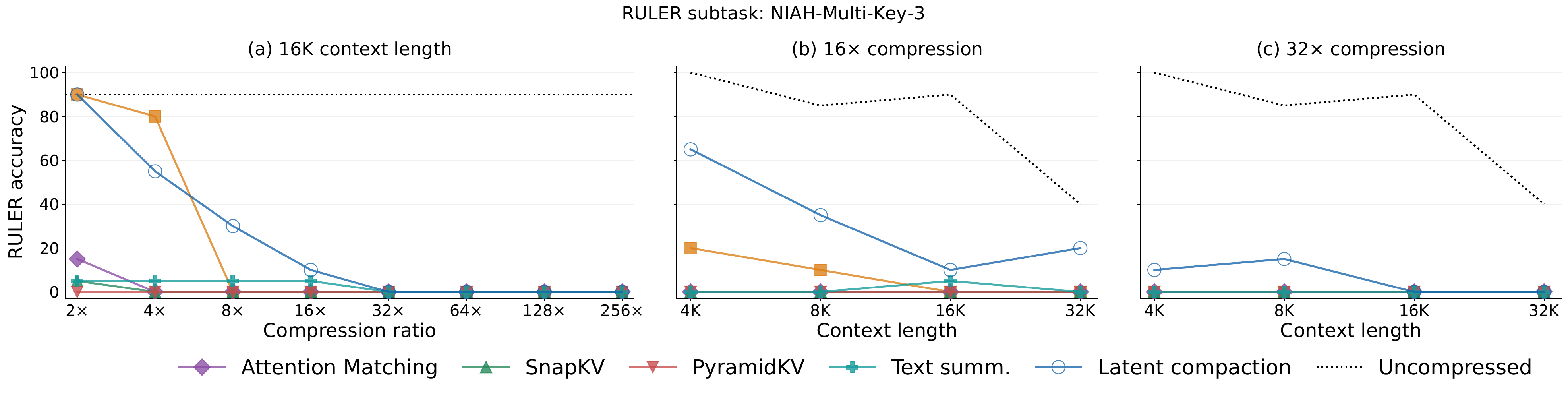}
  \caption{RULER NIAH-Multi-Key-3: accuracy versus compression ratio at 16K context (left), and versus context length at $16\times$ (middle) and $32\times$ compression (right). }
  \label{fig:ruler-niah-multikey-3}
\end{figure*}

\begin{figure*}[t]
  \centering
  \includegraphics[width=\textwidth]{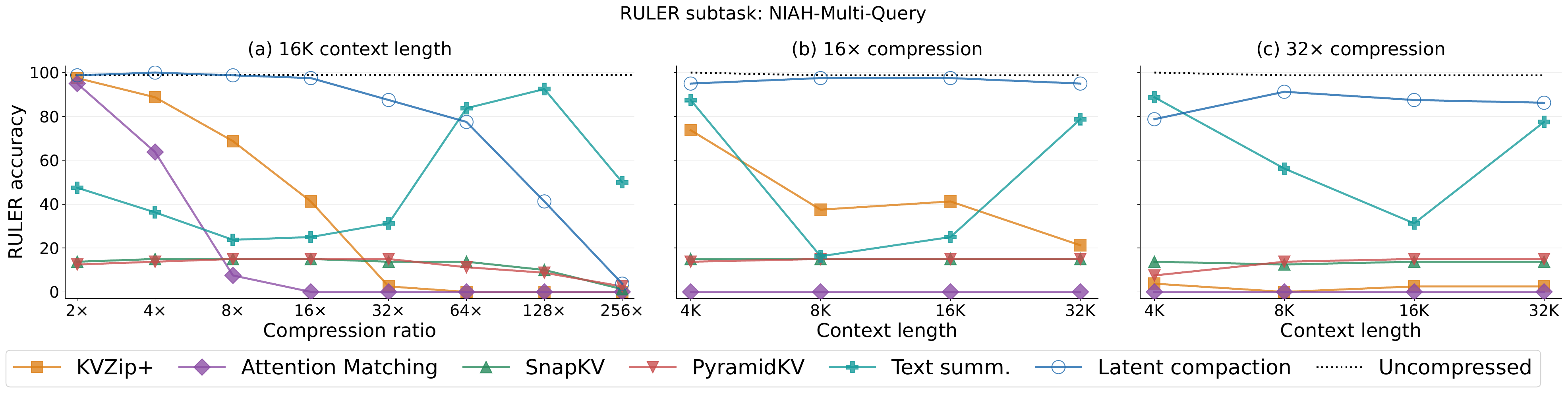}
  \caption{RULER NIAH-Multi-Query: accuracy versus compression ratio at 16K context (left), and versus context length at $16\times$ (middle) and $32\times$ compression (right). }
  \label{fig:ruler-niah-multiquery}
\end{figure*}

\begin{figure*}[t]
  \centering
  \includegraphics[width=\textwidth]{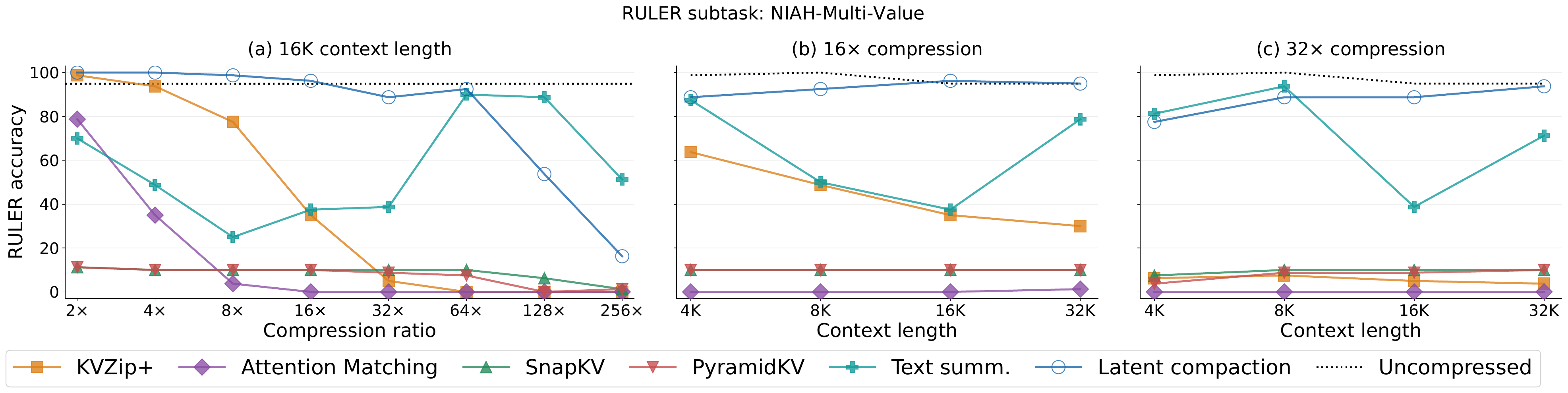}
  \caption{RULER NIAH-Multi-Value: accuracy versus compression ratio at 16K context (left), and versus context length at $16\times$ (middle) and $32\times$ compression (right). }
  \label{fig:ruler-niah-multivalue}
\end{figure*}

\begin{figure*}[t]
  \centering
  \includegraphics[width=\textwidth]{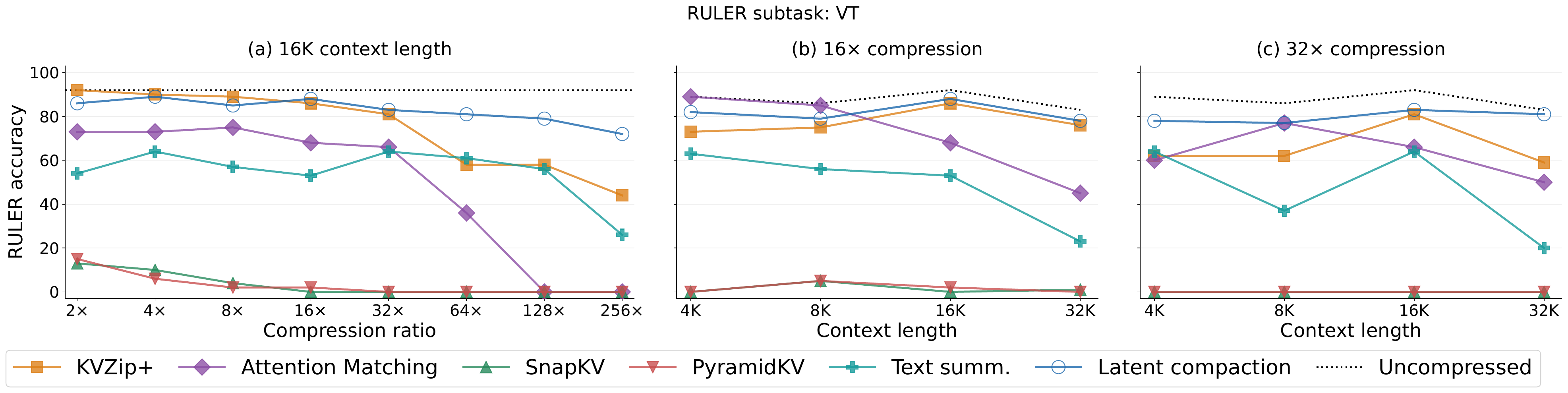}
  \caption{RULER VT: accuracy versus compression ratio at 16K context (left), and versus context length at $16\times$ (middle) and $32\times$ compression (right). }
  \label{fig:ruler-vt}
\end{figure*}

\begin{figure*}[t]
  \centering
  \includegraphics[width=\textwidth]{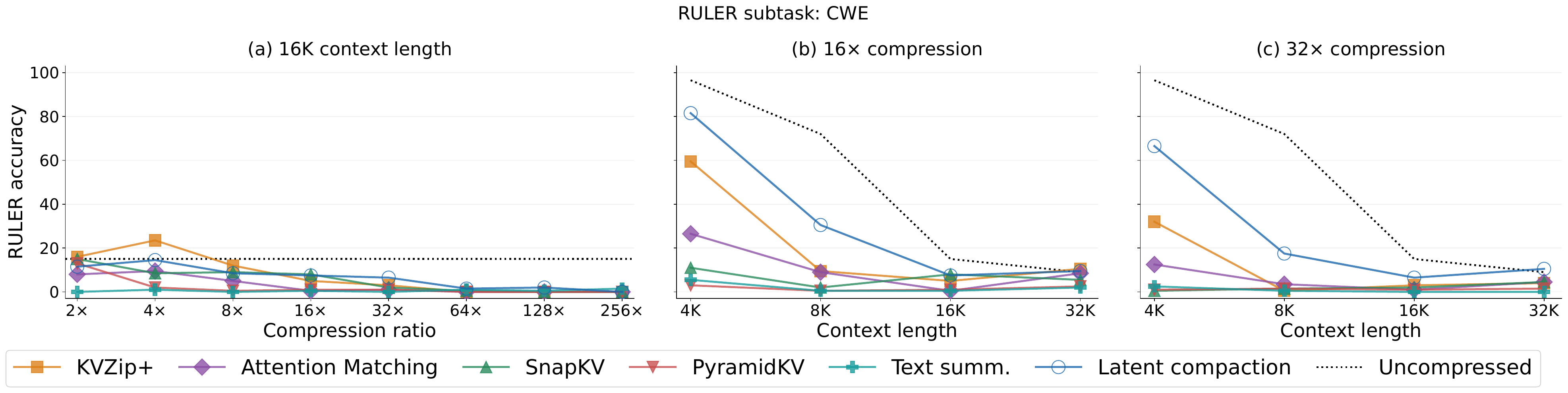}
  \caption{RULER CWE: accuracy versus compression ratio at 16K context (left), and versus context length at $16\times$ (middle) and $32\times$ compression (right). }
  \label{fig:ruler-cwe}
\end{figure*}

\begin{figure*}[t]
  \centering
  \includegraphics[width=\textwidth]{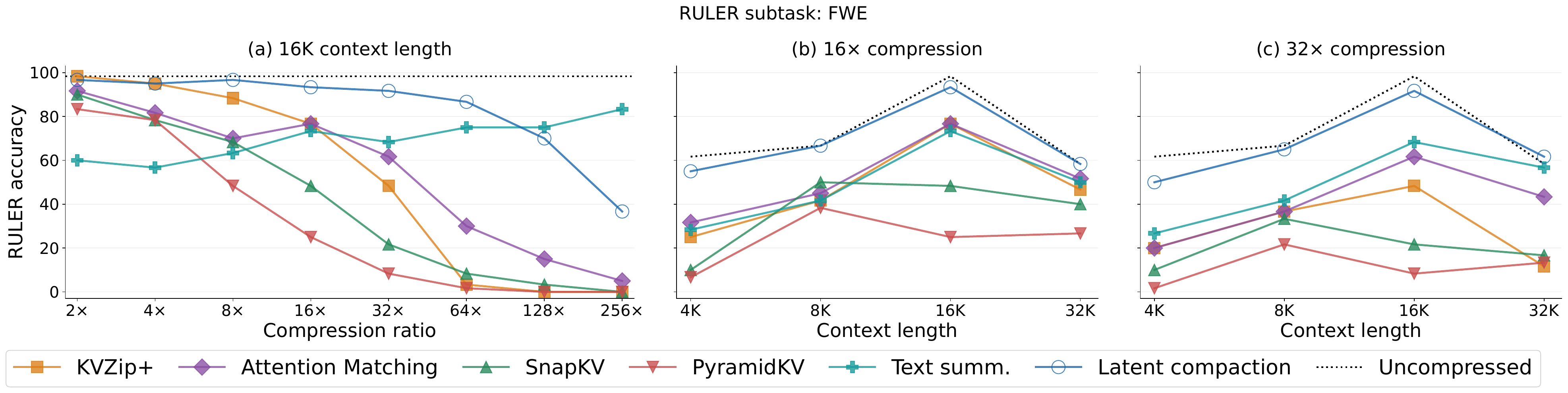}
  \caption{RULER FWE: accuracy versus compression ratio at 16K context (left), and versus context length at $16\times$ (middle) and $32\times$ compression (right). }
  \label{fig:ruler-fwe}
\end{figure*}

\begin{figure*}[t]
  \centering
  \includegraphics[width=\textwidth]{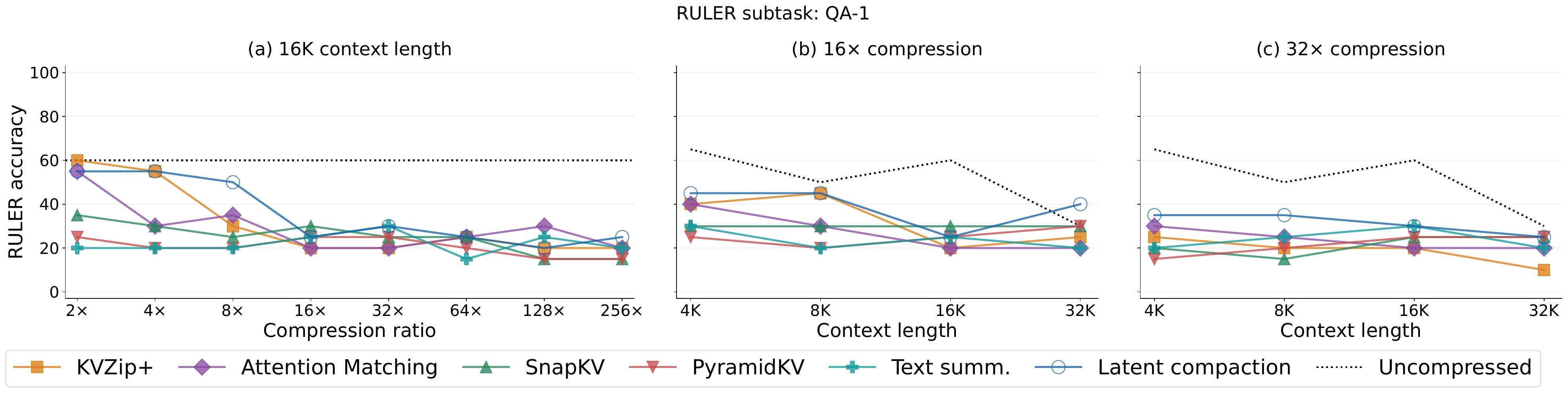}
  \caption{RULER QA-1: accuracy versus compression ratio at 16K context (left), and versus context length at $16\times$ (middle) and $32\times$ compression (right). }
  \label{fig:ruler-qa-1}
\end{figure*}

\begin{figure*}[t]
  \centering
  \includegraphics[width=\textwidth]{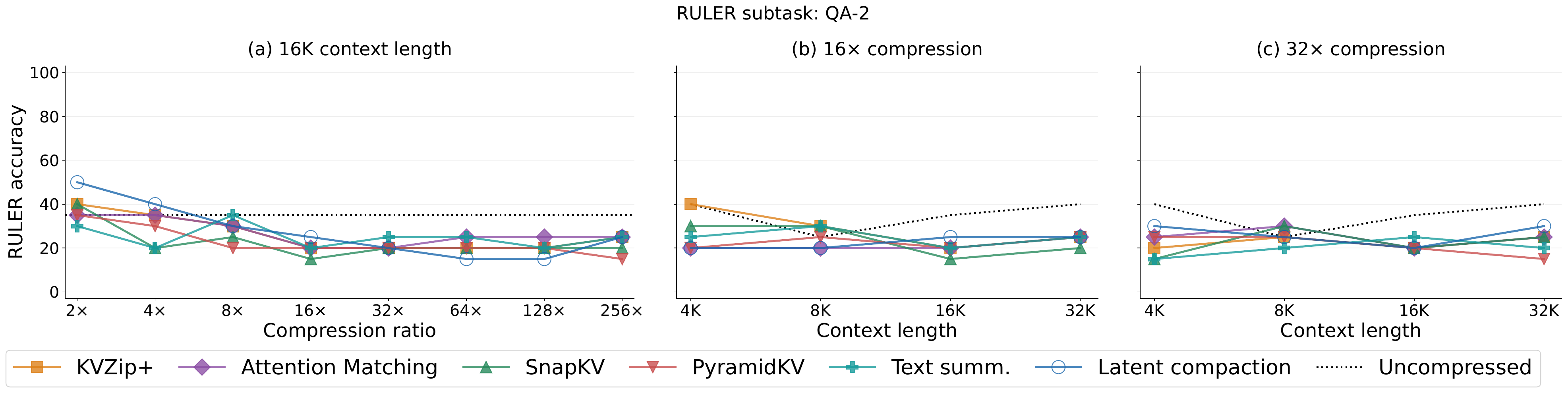}
  \caption{RULER QA-2: accuracy versus compression ratio at 16K context (left), and versus context length at $16\times$ (middle) and $32\times$ compression (right). }
  \label{fig:ruler-qa-2}
\end{figure*}

The aggregate RULER scores can conceal differences among task families. Figures~\ref{fig:ruler-niah-single-1}--\ref{fig:ruler-qa-2} show the results of Latent Compaction and baseline methods on each subtask in the RULER benchmark. Together, these task-level results make clear how accuracy changes with compression ratio at 16K context and with context length at $16\times$ and $32\times$ compression.

\end{document}